# Regression-based Deep-Learning predicts molecular biomarkers from pathology slides

Omar S. M. El Nahhas (1), Chiara M. L. Loeffler (1,2), Zunamys I. Carrero (1), Marko van Treeck (1), Fiona R. Kolbinger (1, 3), Katherine J. Hewitt (1), Hannah S. Muti (1,3), Mara Graziani (4,5), Qinghe Zeng (6), Julien Calderaro (7), Nadina Ortiz-Brüchle (8, 9), Tanwei Yuan (10), Michael Hoffmeister (10), Hermann Brenner (10, 11, 12), Alexander Brobeil (13, 14), Jorge S. Reis-Filho (15), Jakob Nikolas Kather (1, 2, 16, 17)

(1) Else Kroener Fresenius Center for Digital Health, Medical Faculty Carl Gustav Carus, TUD Dresden University of Technology, Germany
(2) Department of Medicine 1, University Hospital and Faculty of Medicine Carl Gustav Carus, TUD Dresden University of Technology, Germany
(3) Department of Visceral, Thoracic and Vascular Surgery, University Hospital and Faculty of Medicine Carl Gustav Carus, TUD Dresden University of Technology, Germany
(4) University of Applied Sciences of Western Switzerland (HES-SO Valais), Rue du Technopole 3, Sierre, 3960, Valais, Switzerland
(5) IBM Research Europe, 8803, Rüschlikon, Switzerland
(6) Centre d'Histologie, d'Imagerie et de Cytométrie (CHIC), Centre de Recherche des Cordeliers, INSERM, Sorbonne Université, Université Paris Cité, Paris, France; Laboratoire d'Informatique Paris Descartes (LIPADE), Université Paris Cité, Paris, France
(7) Assistance Publique-Hôpitaux de Paris, Département de Pathologie, CHU Henri Mondor, F-94000 Créteil, France; Université Paris-Est Créteil, Faculté de Médecine, Créteil, France
(8) Institute of Pathology, University Hospital RWTH Aachen, Aachen, Germany
(9) Center for Integrated Oncology Aachen Bonn Cologne Duesseldorf (CIO ABCD), Germany
(10) Division of Clinical Epidemiology and Aging Research, German Cancer Research Center (DKFZ), Heidelberg, Germany
(11) Division of Preventive Oncology, German Cancer Research Center (DKFZ) and National Center for Tumor Diseases (NCT), Heidelberg, Germany
(12) German Cancer Consortium (DKTK), German Cancer Research Center (DKFZ), Heidelberg, Germany
(13) Institute of Pathology, University Hospital Heidelberg, 69120 Heidelberg, Germany
(14) Tissue Bank, National Center for Tumor Diseases (NCT), University Hospital Heidelberg, 69120 Heidelberg, Germany
(15) Department of Pathology and Laboratory Medicine, Memorial Sloan Kettering Cancer Center, New York, NY, USA
(16) Pathology & Data Analytics, Leeds Institute of Medical Research at St James's, University of Leeds, Leeds, United Kingdom
(17) Medical Oncology, National Center for Tumor Diseases (NCT), University Hospital Heidelberg, Heidelberg, Germany

## Abstract

Deep Learning (DL) can predict biomarkers from cancer histopathology. Several clinically approved applications use this technology. Most approaches, however, predict categorical labels, whereas biomarkers are often continuous measurements. We hypothesized that regression-based DL outperforms classification-based DL. Therefore, we developed and evaluated a new self-supervised attention-based weakly supervised regression method that predicts continuous biomarkers directly from images in 11,671 patients across nine cancer types. We tested our method for multiple clinically and biologically relevant biomarkers: homologous repair deficiency (HRD) score, a clinically used pan-cancer biomarker, as well as markers of key biological processes in the tumor microenvironment. Using regression significantly enhances the accuracy of biomarker prediction, while also improving the interpretability of the results over classification. In a large cohort of colorectal cancer patients, regression-based prediction scores provide a higher prognostic value than classification-based scores. Our open-source regression approach offers a promising alternative for continuous biomarker analysis in computational pathology.

## Introduction

The collection and pathological examination of tissue specimens is used for accurate diagnosis of patients with malignant tumors, providing information related to histology grade, subtype, stage and other tumor biomarkers. Digital pathology describes the computational analysis of tissue specimen samples in the form of whole slide images (WSI). Numerous studies have shown that alterations in individual genes[1–3], microsatellite instability[4–6], and the expression of individual genes[7] or expression patterns of groups of genes[8,9] can be predicted directly from WSI. This research area has also enabled genetic changes to be correlated with morphological patterns (i.e. genotypic-phenotypic correlations)[10], which facilitates the prediction of patient outcome.[11] Consistent with their clinical application, several of these methods have been approved for clinical use by regulatory agencies[12], to the extent that the prediction of biomarkers from pathological diagnostic workflows based on deep learning (DL) is becoming increasingly relevant, not only in the research setting, but also as a de facto clinical application.[2,12,13]

The prediction of genotypic-phenotypic correlations, which involves predicting genetic biomarkers from WSIs, is a weakly supervised problem in DL. To accomplish this task, a DL model correlates phenotypic features from WSIs with a single ground truth obtained from molecular genetic sequencing of tumor tissue at the patient level. Nevertheless, as these WSI are of gigapixel resolution, neural network processing requires breaking them into smaller regions referred to as tiles or patches. These regions may, however, contain less relevant tissues such as connective tissue or fat, which might not contribute to biomarker predictability.[14] To address this issue, attention-based multiple instance learning (attMIL) is the predominant technical approach that is currently used.[15–18] To implement this strategy, feature vectors are first extracted from pre-processed tiles. These vectors are then aggregated by a multi-layer perceptron with an attention component, allowing for a patient-level prediction of the WSI.

Despite the current attMIL approach yielding a high accuracy for biomarker prediction from WSIs[15,19,20], almost all published approaches are limited to classification problems with categorical values (e.g. presence or absence of a genetic alteration).[1–3,8,11,21,22] Nonetheless, the ground truth of many biomarkers is available as continuous values, which are then binarized prior to being utilized as ground-truth for DL. This is true for whole-genome duplications, copy number alterations, homologous recombination deficiency (HRD), gene expression values, protein abundance, and many other measurements. Studies that pursue regression analysis of continuous values often opt for dichotomisation or custom thresholds for categorization. For example, prior to modeling, Fu et al. utilized a LASSO approach for the classification of continuous chromosome data into three classes.[10] Schmauch et al. trained a regression model to predict continuous biomarkers and subsequently used percentile thresholds for the evaluation of the models through a categorical representation.[7]

However, binarization or dichotomization of these values results in information loss[23], which presumably limits the performance of DL systems predicting these biomarkers from pathology slides. Alternatively, a more suitable approach to classification in histopathological WSI analysis would be regression. Regression[24] is a modeling approach used to investigate the relationship between variables, such as morphological features from a WSI, and continuous numerical values, such as genetic biomarkers. To date, there is a paucity of data exploring this approach. A recent study by Graziani et al. presented a novel approach to predict continuous values from pathological images[25], yet their regression network was not systematically compared and required more extensive validation with respect to the more-explored classification approach.

In this study, we systematically compared classification- and regression-based approaches for prediction of continuous biomarkers across multiple cancer types. We hypothesized that regression outperforms classification in weakly supervised analyses of pathology hematoxylin-and-eosin (H&E)-stained WSIs for biomarker predictability, model interpretability and prognostic capability. In addition to various tumor entities, our work also explores several clinically relevant biomarkers represented as continuous numerical values. As a result, we developed a new contrastively-clustered attention-based multiple instance learning (CAMIL) regression approach, which combines self-supervised learning (SSL) with attMIL, and systematically compared it with the CAMIL classification approach, and the regression method proposed by Graziani et al.[25] The evaluation and application of regression versus classification on multiple datasets, organs and biomarkers fills a gap in the computational pathology literature.

# Results

### Regression predicts HRD from histology

We developed a new regression-based DL approach which combines a feature extractor trained by SSL[26] and an attMIL[14] model (**Fig. 1A-C**), referred to as contrastively-clustered attention-based multiple instance learning (CAMIL) regression. We tested the abilities of this approach for prediction of HRD directly from pathology images. We chose HRD because it is a pan-cancer biomarker that is measured as a continuous score, but can be binarized at a clinically validated cutoff. We used the The Cancer Genome Atlas (TCGA) cohorts for breast cancer (BRCA), colorectal cancer (CRC), glioblastoma (GBM), lung adenocarcinoma (LUAD), lung squamous cell carcinoma (LUSC), pancreatic adenocarcinoma (PAAD), and endometrial cancer (UCEC) to train a regression DL model for each cancer type and evaluated their performance by cross-validation (**Fig. 1D**). To mitigate batch effects, which are problematic in the TCGA cohort, we used site-aware cross-validation splits[27]. We found that our CAMIL regression models were able to predict HRD status with AUROCs above 0.70 in 6 out of 7 tested cancer types. The area under the receiver operating characteristic (AUROC) with 95% confidence interval (CI) were 0.78 (± 0.02) in BRCA, 0.76 (± 0.12) in CRC, 0.75 (± 0.40) in GBM, 0.72 (± 0.06) in PAAD, 0.72 (± 0.05) in LUAD, 0.57 (± 0.05) in LUSC, and 0.82 (± 0.03) in UCEC (**Fig. 2A, Suppl. Table 1**). We validated the models on CPTAC, a set of external validation cohorts, in which images and HRD status were available for LUSC, LUAD, PAAD, UCEC. In these cohorts, the model achieved even higher AUROCs, reaching 0.68 (± 0.04) in PAAD, 0.81 (± 0.03) in LUAD, and 0.96 (± 0.01) in UCEC. The lowest AUROC was 0.62 (± 0.06) in LUSC. Together, these data show that regression-based DL can predict HRD status from pathology images alone.

## Regression outperforms the state-of-the-art classification-based approach

We compared the performance of our new DL approach, CAMIL regression, against two state-of-the-art approaches: the Graziani et al. regression method[25] and the CAMIL classification method. In order to compare classification with regression, we chose the AUROC as an evaluation metric. In the site-aware-split test set of the TCGA cohort, CAMIL regression outperformed both of the previous approaches in HRD prediction in all 7 of the tested cancer types (**Fig. 2A, Suppl. Table 1**). In 5 out of 7 cancer types, an ANOVA test showed that the difference in mean AUROCs was statistically significant with $p<0.05$ (**Suppl. Table 2 and 3**). In TCGA-LUSC, all three methods performed equally poorly, reaching AUROCs of 0.57 (± 0.05), 0.57 (± 0.04) and 0.57 (± 0.03) for CAMIL regression, Graziani et al. regression, and CAMIL classification, respectively. In the external validation cohorts, all models reached comparable performance (**Suppl. Table 1 and 2**). In the external validation cohorts (**Fig. 2B**), a t-test showed that the mean AUROCs of CAMIL regression were not statistically significantly better than the classification model, whereas the Graziani et al. model outperformed the CAMIL classification model in 1 out of 4 external validation cohorts (**Suppl. Table 3**).

Next, we compared CAMIL regression to Graziani et al.[25] regression by assessing the coefficient of determination $R^2$ of the predicted scores compared to the clinically-derived ground-truth scores. In TCGA, the CAMIL regression model reached higher $R^2$ scores than the Graziani et al.[25] model in all of the 7 selected cohorts (**Suppl. Table 5**). In the CPTAC validation cohort, the CAMIL regression model reached higher $R^2$ scores than the Graziani et al.[25] model in all 4 of the selected cohorts (**Suppl. Table 5**). To determine the reason for our superior performance over Graziani et al.[25] regression, we conducted an ablation study of the CAMIL regression approach. These results revealed that the inferior performance in Graziani et al.[25] approach for predicting clinical biomarkers is mainly due to the standard stochastic gradient descent optimizer, compared to the stochastic gradient descent with adaptive moments optimizer in our CAMIL regression approach (**Suppl. Table 7**). Taken together, these data indicate that the CAMIL regression method outperforms the Graziani et al.[25] regression method and the CAMIL classification method. Consequently, the regression method by Graziani et al.[25] is not further compared to CAMIL regression and classification in subsequent experiments.

Moreover, we investigated additional aspects of model performance which the AUROC does not capture[28]. We compared CAMIL regression to CAMIL classification by quantifying the absolute distance between the medians of the normalized scores for the positive and negative samples (**Fig. 2C-F**). For example, for detection of HRD status in endometrial cancer, the AUROC on the CPTAC test cohort was 0.98 ± 0.02 for CAMIL classification and 0.96 ± 0.01 for CAMIL regression. This difference was not statistically significant ($p = 0.095$). When the distribution of the CAMIL regression model output (**Fig. 2C-F**) was visualized, we found a greater separation of the predicted HRD scores in positive and negative patients compared to the CAMIL classification approach (**Suppl. Table 4**). The absolute distance between the peak of the score distribution between positive and negative patients was higher for CAMIL regression than for CAMIL classification. We further quantified this in all tumor entities and found that in all 7 of the selected TCGA cohorts, this distance was larger in the CAMIL regression, resulting in a greater class separability. In CPTAC, as compared to the classification-based approach, class separability was improved in 2 out of 5 cohorts when using the regression approach. Overall, our CAMIL regression approach improves separation distance of the groups' medians by 378% for the test set of the TCGA training cohort, and 19% for the external CPTAC test cohort (**Suppl. Table 4**).

## Regression predicts key biological process biomarkers from histology

Having shown that our CAMIL regression method can predict HRD from histology WSIs, we expanded our experiments to additional biomarkers. We investigated biomarkers related to the three key components of solid tumors: tumor cells, stroma, and immune cells. For tumor cells, we aimed to predict proliferation, as measured by an RNA expression signature[29]. For stroma, we aimed to predict stromal fraction (SF), as assessed via DNA methylation analysis[29]. For immune cells, we investigated the tumor infiltrating lymphocytes regional fraction (TIL RF), the leukocyte fraction (LF), and the lymphocyte infiltration signature score (LISS)[29]. We found that our CAMIL regression method was able to predict all of these five biomarkers with high AUROCs across cancer types in the TCGA cohort (**Suppl. Table 9**). For example, in breast cancer, the AUROCs for these five biomarkers were 0.88 (± 0.02) in TIL RF, 0.83 (± 0.05) in proliferation, 0.81 (± 0.03) in leukocyte fraction, 0.80 (± 0.03) in LISS and 0.80 (± 0.03) in stromal fraction. In colorectal cancer, these AUROCs were 0.79 (± 0.07), 0.59 (± 0.12), 0.76 (± 0.06), 0.70 (± 0.01) and 0.77 (± 0.04), respectively. In all other cancer types, mean AUROCs of above 0.70 were reached (**Suppl. Table 9**). These findings show that the regression-based DL model can be trained to predict tumor cell proliferation, stromal fraction and immune-cell-related biomarkers from H&E histopathology.

We compared this to the state-of-the-art CAMIL classification approach using the AUROC with 95%CI as evaluation metric. Using site-aware splits, our proposed CAMIL regression approach outperformed CAMIL classification in 8 out of 34 instances, with the remainder of cases having equal performance for the classification and regression approach (**Fig. 3B**). Regression outperformed classification in TCGA-BRCA in two targets, LF (0.80 ± 0.02, p<0.0001) and LISS (0.80 ± 0.03, p<0.0001). In TCGA-CRC, the performance between regression and classification was equal for all five targets. In TCGA-LIHC, regression outperformed classification in LISS (0.70 ± 0.01, p < 0.001). In TCGA-LUAD, regression outperformed classification in proliferation (0.84 ± 0.04, p < 0.0001). In TCGA-LUSC, regression outperformed classification in TIL RF (0.88 ± 0.04, p < 0.0001). In TCGA-STAD, regression outperformed classification in proliferation (0.87 ± 0.07, p = 0.06), but did not reach a statistically significant AUROC in either classification or regression (p > 0.05). In TCGA-UCEC, regression outperformed classification in the two lymphocyte-based targets, TIL RF (0.82 ± 0.04, p < 0.0001) and LISS (0.73 ± 0.06, p < 0.001). These findings collectively demonstrate that utilizing the CAMIL regression approach leads to an average 4% increase in the AUROCs, as compared to employing the CAMIL classification approach for the same task of predicting key biological process biomarkers from histology.

**Regression improves interpretability of biomarker predictions from histology**
Next, we investigated the interpretability of the CAMIL classification model compared to the CAMIL regression model. We evaluated the biological plausibility of spatial prediction heatmaps obtained by deploying the regression model and the classification model on tumors in the site-aware split test set of the TCGA cohort. We used the models trained to predict the LISS in breast cancer. Although the LISS is only available as a weak label (one score per WSI), a good model should be able to highlight regions which are associated with the LISS, and these regions should contain lymphocytes. Indeed, we saw that both the classification model and the regression model placed their attention on lymphocyte-rich regions (**Fig. 3C-0**). In the evaluated WSIs, however, the LISS regression model yielded a sharper delineation of lymphocyte-rich regions and placed less attention on areas where histologic features are less relevant. Contrastingly, the LISS classification model demonstrates relatively less confidence in areas with a dense lymphocyte population compared to the regression model, as indicated by a lower attention score (**Fig. 3C-1**). The classification model assigns importance to regions without any presumed clinical relevance, as evidenced by the fact that the model highlighted the tissue edge which lacks high density lymphocytes regions (**Fig. 3C-2**). We quantified these findings by a blinded interpretability review of 42 attention heatmaps from the classification and regression models by KJH, a pathology resident. Based on the expert review, the CAMIL regression approach produced the most interpretable attention heatmaps in 34 out of 42 cases. In 6 out of 42 cases, the CAMIL classification approach was more interpretable. Similar interpretability between the CAMIL classification and regression approaches was observed in 2 out of 42 cases. Hence, CAMIL regression outperforms CAMIL classification in interpretability in 81% of cases as observed in a blinded review. Taken together, these data demonstrate that the regression approach gives a statistically significantly better AUROC for the investigated biomarkers ($p < 0.05$; **Suppl. Table 11)**, and a markedly improved interpretability, compared to the classification approach.

**Regression-based biomarkers improve survival prediction in colorectal cancer**
Biological processes of tumor cell proliferation, deposition of stromal components, and infiltration by lymphocytes are biologically relevant during tumorigenesis and progression, and are known to be related to clinical outcome.[30,31] Thus, prediction of lymphocytic infiltration from H&E pathology slides should be relevant for prognostication. We investigated this in a large cohort of 2,297 patients with colorectal cancer from the Darmkrebs: Chancen der Verhütung durch Screening (DACHS) study, for which H&E WSIs and long-term (10 years) follow-up data were available for overall survival (**Suppl. Table 15)**.

First, we deployed the CAMIL classification models that were trained on colorectal cancer patients in TCGA, which obtained similar AUROCs in all biomarkers (**Fig. 3B**). We deployed these models on WSIs from patients enrolled in DACHS, obtaining a binarized prediction label for each patient. We then assessed the prognostic impact of this predicted label with univariate and multivariate Cox Proportional Hazard models for overall survival (**Fig. 4A and 4B**), yielding hazard ratios (HR). We found that the classification models reached significant risk-group stratification in 3 out of 5 biomarkers (**Fig. 4A, Suppl. Table 12**): leukocyte fraction (HR=0.74, $p < 0.0001$), LISS (HR=0.74, $p < 0.0001$), and stromal fraction (HR=0.77, $p < 0.0001$). These hazard ratios represent only a modest predictability of survival. In the multivariate survival model (**Fig. 4B, Suppl. Table 13**), the classification models show significant prognostic capabilities in only 2 out of 5 biomarkers: leukocyte fraction (HR=0.83, $p = 0.0394$) and LISS (HR=0.82, $p = 0.0265$).

When we repeated the procedure with continuous scores obtained from the CAMIL regression models, we found that the regression models markedly improved the survival prediction. The regression model reached significant risk-group stratification in 3 out of 5 biomarkers (**Fig. 4A**): leukocyte fraction (HR=0.18, $p < 0.01$), LISS (HR=0.03, $p < 0.0001$) and TIL regional fraction (HR=0.21, $p < 0.01$). This effect was also observed when the scores obtained from the CAMIL regression model were binarized at the median before using them as an input for the univariate Cox Proportional Hazard model (**Suppl. Table 14)**, showing consistent risk-group stratification superiority for regression-based biomarkers. For the multivariate survival model (**Fig. 4B, Suppl. Table 13**), the regression models show significant prognostic capabilities in the same 2 biomarkers: leukocyte fraction (HR=0.20, $p < 0.01$) and LISS (HR=0.14, $p < 0.01$). Again, the HR for regression are significantly further away from non-significance (HR=1) with non-overlapping 95%CI compared to the classification models. Similar observations were made for the models trained on breast cancer patients from TCGA and deployed on colorectal cancer patients from DACHS, corroborating the improved generalizability of regression on biomarkers across different cancer types (**Fig. 4C and 4D**).

Taken together, these data demonstrate that by training models on biologically relevant biomarkers with weakly supervised learning, the resultant regression models are better predictors of survival than their classification counterpart. Therefore, regression models enhance the use of weakly supervised learning to build DL systems of potential clinical utility.

# Discussion

Since 2018, the field of digital pathology has rapidly expanded to include the development of tools for predicting molecular biomarkers from routine tumor pathology sections, which has led to the development of clinically approved products. Traditional DL methods have limited the analysis of many biomarkers, including HRD and gene expression signatures, which are continuous values, by categorizing them into discrete classes. Our study provides direct evidence that novel regression networks, such as the CAMIL regression method described in this study, which builds on recent work using attention-based multiple instance learning and self-supervised pre-training of the feature extractor[18,20,26], outperforms traditional classification networks in predicting these biomarkers. This approach unlocks a key clinical application area for pathology-based biomarker prediction.

Our proposed CAMIL regression approach has shown promising results in improving the accuracy and separability of biomarker predictions compared to CAMIL classification. This improvement is particularly noticeable for biomarkers that have a clinically established threshold for categorization, such as HRD. Similar improvements are observed for biomarkers that do not have any clinically relevant cut-off point and would traditionally necessitate dichotomization for analysis, such as immune biomarkers. In addition, our CAMIL regression approach demonstrates better generalization capabilities than the regression approach by Graziani et al.[25], as seen in the external test cohort. We identified that the optimizer used in Graziani et al.[25] predominantly caused the regression model to converge to the mean, which explains the observed difference.

Furthermore, our study highlights the advantages of regression-based biomarker prediction over classification-based prediction in terms of interpretability. We demonstrated that, for tumor infiltrating lymphocytes, attention heatmaps generated through regression were preferred in 81% of cases for their interpretability compared to those generated through classification. Regression also resulted in an improvement in survival prediction based on immunologic biomarkers, as it allowed for more effective stratification of risk groups for overall survival compared to classification models. The biomarkers were deliberately chosen on the basis of their prognostic capabilities[32–35], and are better reflected by the tumor morphology analysis through the CAMIL regression approach as compared to the CAMIL classification approach.

This study has several limitations. The experiments were limited to a select number of tumors and clinical targets, and not all analyzed clinical targets had an external test set with the same clinical information available. This resulted in meta-external test sets through site-aware splits, and blind deployments on an external cohort. Additionally, none of the hyperparameters of the trained models were optimized. Further research could expand the analysis to a wider variety of cancers and clinical targets, while also exploring potential pitfalls of regression in computational pathology. The approaches described here, however, provide a proof-of-principle for the use of regression-based attMIL systems and their potential impact for the inference of biomarkers and prediction of outcomes from histologic WSIs, expanding the repertoire of applications of DL in precision medicine.

# Materials and Methods

## Ethics statement

We examined anonymized patient samples from several academic institutions in this investigation. This analysis has been approved by the ethical boards at DACHS. CPTAC and TCGA did not require formal ethics approval for a retrospective study of anonymised samples. The overall analysis was approved by the Ethics commission of the Medical Faculty of the Technical University Dresden (BO-EK-444102022).

Image Data and Cohorts

A total of 11,671 raw WSIs were scanned by an Aperio ScanSlide scanner and pre-processed in this study. Two types of clinical targets were analyzed to observe the performance of the classification and regression models: 1) continuous variables with a known clinically relevant cut-off for categorization, and 2) continuous variables with unknown clinically relevant cut-offs, thus requiring categorization by splitting at the median. These categories of targets were chosen due to theory mentioning the loss of information by splitting at the median[23], but does not mention the loss of information when utilizing clinically relevant cut-offs before training the model.

The target with a clinically relevant cut-off is homologous recombination deficiency (HRD) (**Suppl. Table 16**), a clinically relevant biomarker in solid tumor types, such as breast cancer. One way to calculate HRD is by adding up the three subscores, Loss of Heterozygosity (LOH), Telomeric Allelic Imbalance (TAI) and large-scale state transitions (LST), giving us a continuous value ranging from 0 to 103 in the training sets. A clinically relevant cut-off point of HRD >= 42 was used to binarize the continuous score[36].

The targets without a known clinically relevant cut-off point are biological process biomarkers (**Suppl. Table 17**), which are interesting to analyze due to their prominent role in immunotherapy outcome prediction[29,37,38]: Stromal Fraction (SF) with range [0, 0.92] and leukocyte fraction (LF) with range [0, 0.96] as assessed via DNA methylation analysis, lymphocyte infiltrating signature score (LISS) with range [-3.49, 4.17] and proliferation (Prolif.) with range [-2.86, 1.59], as measured by RNA expression data and tumor infiltrating lymphocytes regional fraction (TIL RF) with range [0, 63.65], quantified using a DL based classification. For TCGA-LIHC, there was no data available for TIL regional fraction, leading to an analysis of 5 targets in 7 cancer types with 5-fold cross-validation, resulting in (35-1)*5 models for each modeling type, of which the AUROC ± 95%CI of the 5 folds per target and tumor type was reported.

Model description

The entire image processing pipeline, from whole-slide image (WSI) to patient-level predictions, consisted of three main steps: 1) image pre-processing, 2) feature extraction, 3a) classification-based attention attMIL and 3b) regression-based attMIL for score aggregation resulting in patient-level predictions (**Fig. 1A and 1B**).

All WSI in the experiments were tessellated into image patches at a resolution of 224 by 224 pixels with an edge length of 256 µm, resulting in a Microns Per Pixel (MPP) value of approximately 1.14. After tessellation, every image patch underwent a rejection filter using the Canny edge detection method[39], removing blurry patches and the white background of the image when two or less edges were detected in the patches. The remaining patches were color-normalized in order to reduce the H&E-staining variance across patient cohorts according to the Macenko spectral matching technique[40], with a prior added step of brightness standardization. For pre-processing, our end-to-end WSI pre-processing pipeline was utilized. The target image used to define the color distribution was uploaded to the GitHub repository.

Every pre-processed image patch was turned into a 2048 feature vector through inference of a ImageNet-weighted ResNet50-based self-supervised contrastive clustering model fine-tuned on 32,000 WSIs from different cancer types; RetCCL[26]. The feature extraction resulted in an *(n x 2048)* feature matrix per patient, where n is the number of *(224 x 224 pixels)* pre-processed image patches.

Experimental setup and implementation details

For the experiments, 5-fold cross-validation on patient-level with site-aware splits was performed to train the models. Site-aware splits ensure that patients are stratified and grouped by the hospital the WSI originated from, creating a stratified random 80-20 split which forces all patients from the same hospital to exist in either the training and internal validation set, or the internal test set, while retaining ground-truth class distributions. Specifically, in The Cancer Genome Atlas (TCGA), site-specific histological features were shown to be present in the WSI, causing biased evaluations in the model when not accounted for accordingly during the training procedure[27]. The basis for the weakly supervised classification and regression was adapted from the attention-based multiple instance learning (attMIL) method by Ilse et al [41]. Our proposed model used Balanced MSE[42] as a loss function to account for the natural class imbalance in clinical settings, as well as the Adam optimizer[43] and an attention component followed by a MLP head[41] which was trained for 25 epochs. The dropout layer was removed, due to loss of performance in regression in tabular data settings[44]. The attMIL variant in our proposed CAMIL regression differs from Ilse et al. by swapping their feature extractor with a pre-trained ResNet50 with ImageNet weights, fine-tuned on 32,000 histopathology images in a self-supervised manner using contrastive clustering shown to yield significantly better results on WSI image analysis[26]. Moreover, the classification head consisting of a fully-connected (FC) layer and sigmoid operation was swapped with custom heads to allow for classification and regression tasks to be performed. The attention component was not altered.

To evaluate the relative supremacy between classification and regression, first, the CAMIL regression method was compared with 1) the regression method from Graziani et al. and 2) the CAMIL classification method on the continuous HRD score and clinically-relevant binarized HRD score, respectively. Then, CAMIL regression was compared to CAMIL classification on continuous biomarkers related to biological processes with no known clinically-relevant cut-off points, where the median score per target was used for binarizing. Moreover, an expert review by a pathology resident was conducted on attention heatmaps produced by CAMIL classification and CAMIL regression to determine which method yielded the most interpretable heatmaps. Finally, the prognostic capabilities of CAMIL regression versus CAMIL classification was evaluated on an external data cohort DACHS-CRC by predicting survival of groups stratified by the models which were trained on the same biological process biomarkers and extracted features. For the HRD scores, the models were trained on TCGA-BRCA, TCGA-CRC, TCGA-GBM, TCGA-LUAD, TCGA-LUSC, TCGA-PAAD, TCGA-UCEC and externally validated on CPTAC-LUAD, CPTAC-LSCC, CPTAC-PDA and CPTAC-UCEC. For the biological process biomarkers, the models were trained on TCGA-BRCA, TCGA-CRC, TCGA-LUAD, TCGA-LUSC, TCGA-LIHC, TCGA-STAD and TCGA-UCEC. Every model that was compared, both regression and classification, consisted of the exact same patients for training, internal validation, internal testing and external testing (**Suppl. Table 16 and 17**).

For the regression method from Graziani et al., we introduced the self-supervised component as feature extractor[26] followed by embedding-level attention aggregation, instead of the ImageNet weighted ResNet18 backbone followed by patch-level attention aggregation in the original study by Graziani et al.[25] As it was shown that the self-supervised backbone increases performance and generalizability compared to an ImageNet weighted architecture as backbone[26], we added the self-supervised component in order to compare the regression heads in isolation. The commonalities between the models are the learning rate (1.00E-04), weight decay (1.00E-02), patience (12 epochs), the attention component[41] and the fit-one-cycle learning rate scheduling policy[45]. The differences of the models' hyperparameters and optimization strategies (**Suppl. Table 6**) of Graziani et al. and our CAMIL regression model were broken down in an ablation study to find the reason for the performance differences of the regression heads.

Statistics and endpoints

The classification and regression method were made comparable in a similar dimension by utilizing the area under the receiver operating characteristic (AUROC) metric. For the definition of the binarized groups required for the AUROCs, the clinically-relevant cut-off for HRD was used, while for the biological process biomarkers, the continuous targets were split at the median. The prediction scores of the classification model [0-1] and the predictions of the regression models $(-\infty, \infty)$ were used as continuous score for all the possible thresholds of the AUROC.[46] By utilizing this approach, it was possible to test which type of model, when provided with the same ground-truth binarized label, had the least overlap between the predicted score distributions for different groups. This, in turn, resulted in achieving the highest AUROC. However, the AUROC measures only the separation of groups' score distributions, but does not account for the distance between the distributions. Therefore, to determine whether there is an increased distance between distributions, the median and interquartile range (IQR) were calculated for the clinically relevant HRD+ and HRD- groups. However, this calculation was not performed for the biological process biomarkers due to the unclear relevance of distance between the dichotomized groups.

To determine statistical significance of the AUROCs, the 95% confidence interval (CI) of the 5 training folds was calculated for each model. In order to identify if the AUROCs of the three compared models (CAMIL classification, regression from Graziani et al., and our proposed CAMIL regression) had a significant difference for the HRD target, the repeated measures ANOVA statistical analysis was performed, which resulted in an F value for each tumor-type the three models were trained on. If the difference between the three models was statistically significant, the dependent one-sided t-test for paired samples statistical analysis was performed in order to determine if CAMIL regression outperformed CAMIL classification, resulting in a t-statistic with 95%CI for every model comparison for every analyzed tumor-type of the internal test set from TCGA. For the external test set, the repeated measures ANOVA is also performed, after which two dependent one-sided t-tests with Bonferroni correction were performed, resulting in two t-statistics with 97.5%CI for every model comparison of every analyzed tumor-type. For the biological process biomarkers' models, a dependent two-sided t-test with 95%CI was performed to test the alternative hypothesis if the 5-fold mean of the CAMIL classification and CAMIL regression AUROCs were significantly different from each other.

To determine the prognostic capabilities of the biological process biomarkers' models, survival prediction analysis is done on an external cohort, DACHS. All 5 models trained through site-aware splits were blindly deployed, of which the mean of the predicted scores were used for further analysis. The univariate (UV) and multivariate (MV) Cox proportional-hazards (PH) regression analysis were independently performed to determine the Hazard Ratio (HR) of the classification and regression models' predictive biomarker. The continuous score from the regression models were used for the Cox PH analyses, as well as the binarized continuous scores to rule out bias in the prognostic capabilities solely through which variant of the continuous score was used. The prognostic capabilities of the classification and regression models were independently analyzed together with three covariates: age (continuous, $\mathbb{R}^+$), sex (binary, 0: female, 1: male) and tumor stage (continuous, $\mathbb{Z} \in [1, 4]$). Thus, one model's scores per target and the three covariates were analyzed for each model independently. Statistical significance of the HR is reached when the 95%CI does not cross a HR=1.

## Visualization and explainability

To compare the separability of CAMIL classification and CAMIL regression models' score distribution for HRD at a similar scale, all values for both models were min-max normalized individually to redistribute every model's score output between [0,1]. To explain the classification and regression CAMIL models' capability of decision-making using clinically relevant features, the attention component from the attMIL model architecture was utilized. The attention heatmaps were created by loading the attMIL model architectures for classification and regression into a fully convolutional equivalent[47] with their respective weights from the training procedure, which allows for a high-resolution attention heatmap, rather than 224x224 patches the model was trained on. By running inference on the WSIs of the patient, the attention layer which resulted from the patient-wise prediction was extracted and used as an overlay on the WSI to indicate hot zones which the model used in decision making. The TCGA-BRCA cohort was chosen for visualization to observe the contrast between equal and superior performance of the regression model compared to the classification model in lymphocyte-based targets. For each target, the classification and regression model were trained, validated and tested on the exact same patient using site-aware splits. The attention heatmaps for the blinded review were generated from patients with the top 42 highest expression of the LISS biomarker from the unseen internal TCGA-BRCA test set through the trained classification and regression models, resulting in 84 heatmaps in total. The models' clinical interpretability was reviewed by a pathologist, choosing the most interpretable attention heatmap for each of the 42 patients.

## Data and Code availability

All source codes are available under an open-source license on GitHub. The pre-processing pipeline is found at https://github.com/KatherLab/end2end-WSI-preprocessing/releases/ tag/v1.0.0-preprocessing, the classification pipeline is found at https://github.com/KatherLab/marugoto/releases/tag/v1.0.0-classification, the regression pipeline is found at https://github.com/KatherLab/marugoto/releases/tag/v1.0.0-regression, and the classification and attention heatmaps are found at https://github.com/KatherLab/highres-WSI-heatmaps/releases/tag/v1.0.0-heatmaps. The slides for TCGA are available at https://portal.gdc.cancer.gov/. The slides for CPTAC are available at https://proteomics.cancer.gov/data-portal. The molecular data for TCGA is available at https://www.cbioportal.org/ and additional biomarker data is available from Thorsson et al.[29]

# Figures

## A Pre-processing

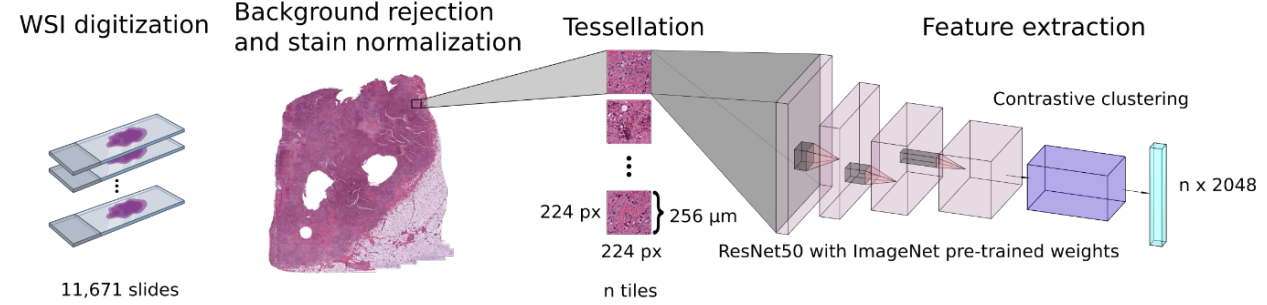

## B Modeling

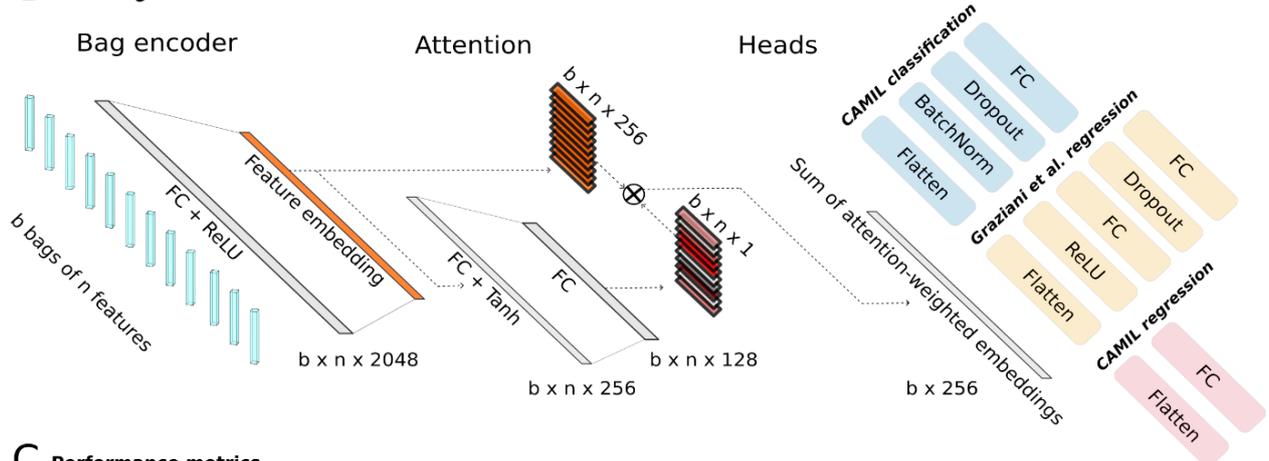

## C Performance metrics

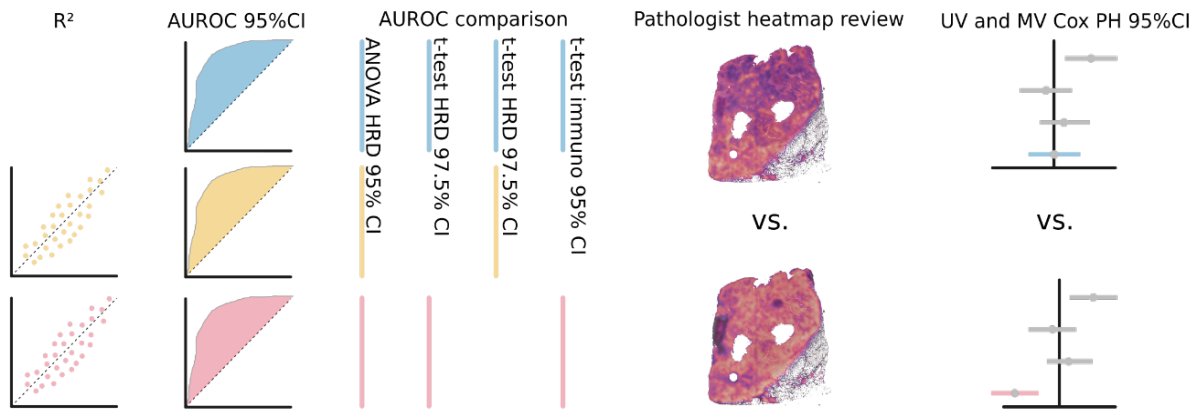

## D Cohorts

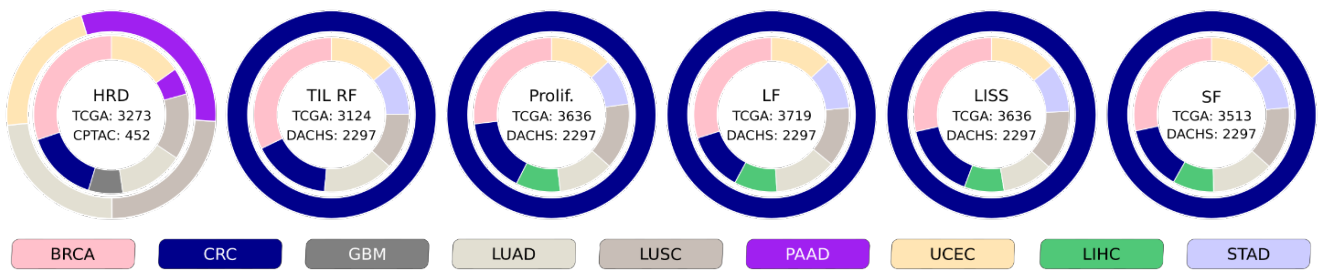

**Figure 1: End-to-end experimental workflow overview with image pre-processing, modeling, performance metrics and used cohorts.** A) The image pre-processing pipeline and tile-level feature extraction by running inference on a ResNet50 with pre-trained ImageNet weights and retrieval contrastive clustering (RetCCL) model for a feature matrix for each patient. B) The modeling architecture using attention-based multiple instance learning (attMIL) on the self-supervised extracted features with three separately trained heads for CAMIL classification, regression as proposed by

Graziani et al., and CAMIL regression as proposed in this study. C) The performance metrics and their respective confidence intervals (CI) to evaluate the performance of the three separately trained heads of the model, including the coefficient of determination ($R^2$) for the regression models, the area under the receiver operating characteristic (AUROC) for all models, analysis of variance (ANOVA) with repeated measures for the homologous recombination deficiency (HRD) and biological process biomarkers, and expert review of attention heatmaps with univariate (UV) and multivariate (MV) Cox proportional-hazard (PH) models for the biological process models. D) The cohorts used for training and external validation represented in the inner- and outer circle, respectively. The training cohorts are from The Cancer Genome Atlas (TCGA) programme for all clinical targets, with the external validation cohorts coming from the Clinical Proteomic Tumor Analysis Consortium (CPTAC) effort and the Darmkrebs: Chancen der Verhütung durch Screening (DACHS) study for the HRD target and the biological process biomarkers, respectively. The biological process biomarkers are tumor infiltrating lymphocytes regional fraction (TIL RF), proliferation (Prolif.), leukocyte fraction (LF), lymphocytes infiltrating signature score (LISS) and stromal fraction (SF). The considered cancer types in this study are breast cancer (BRCA), colorectal cancer (CRC), glioblastoma (GBM), lung adenocarcinoma (LUAD), lung squamous cell cancer (LUSC), pancreas adenocarcinoma (PAAD), endometrial cancer (UCEC), liver hepatocellular carcinoma (LIHC) and stomach cancer (STAD).

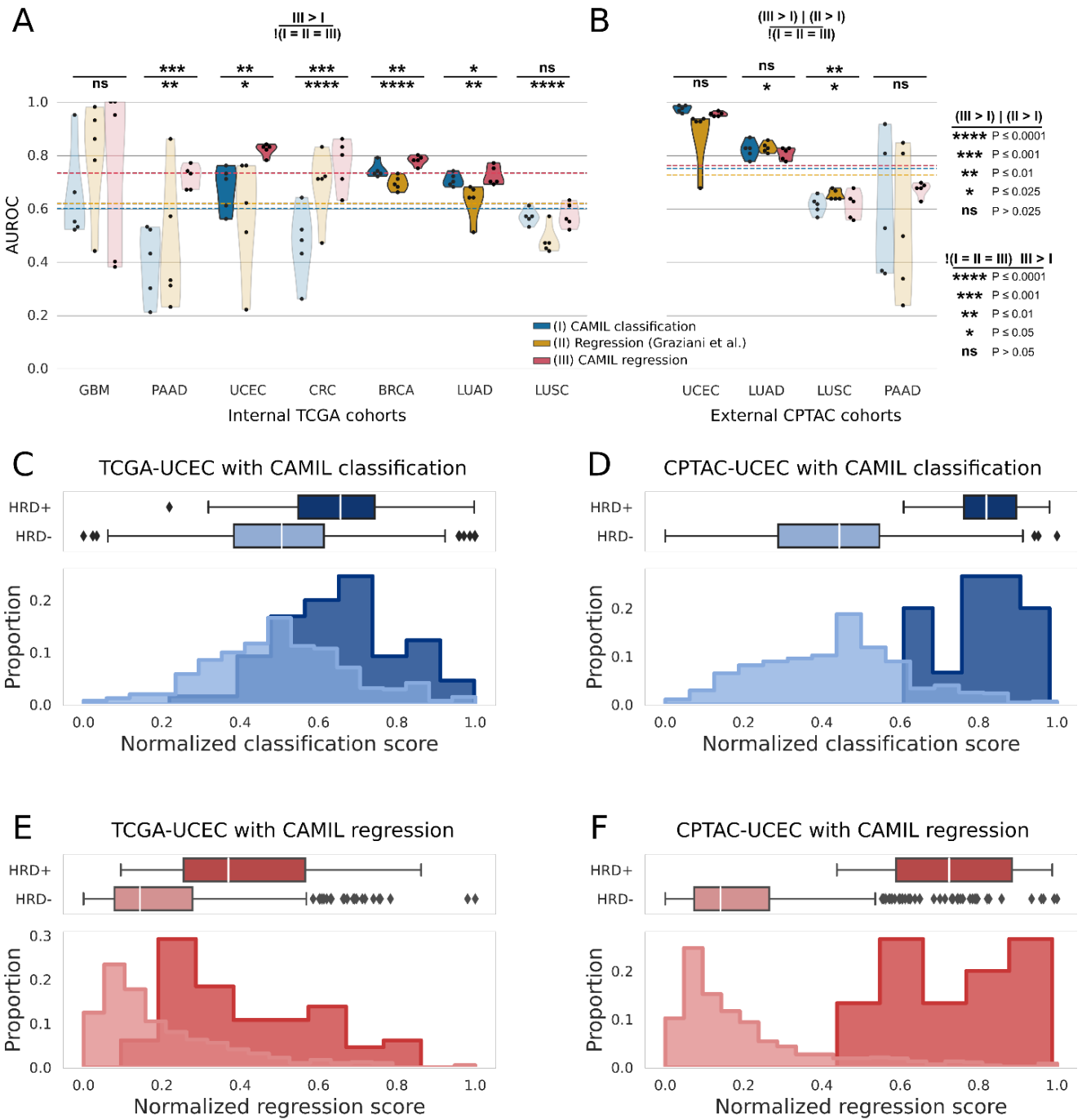

**Figure 2: Performance overview of classification versus regression approaches predicting the homologous recombination deficiency (HRD) score.**
Panel A) and B) show boxplots of area under the receiver operating characteristic (AUROC) values from HRD predictions of this (I) CAMIL classification, (II) regression by Graziani et al. and (III) CAMIL regression on the internal test set from The Cancer Genome Atlas (TCGA) and the external test set from the Clinical Proteomic Tumor Analysis Consortium (CPTAC) effort, respectively. Cancer types include glioblastoma (GBM), pancreas adenocarcinoma (PAAD), endometrial cancer (UCEC), colorectal cancer (CRC), breast cancer (BRCA), lung adenocarcinoma (LUAD) and lung squamous cell cancer (LUSC). Non-significant AUROC values are shown as transparent violin instances, and statistical tests include an analysis of variance with repeated measures displayed at the bottom and dependent one-sided t-tests, with Bonferroni correction for multiple hypothesis testing in the external test set displayed on top. Panel C) and D) show the proportional distribution plot of the normalized classification scores of the internal test set from the CAMIL classification model trained on TCGA-UCEC, and the external test set CPTAC-UCEC, respectively. Panel E) and F) show proportional distribution plot of the normalized regression scores of the internal test set from the CAMIL

regression model trained on TCGA-UCEC, and the external test set CPTAC-UCEC, respectively. In the distribution plots, the ground-truth classes are depicted as a darker shared (HRD+) and lighter shade (HRD-) of the color designated to CAMIL regression and CAMIL classification, respectively.

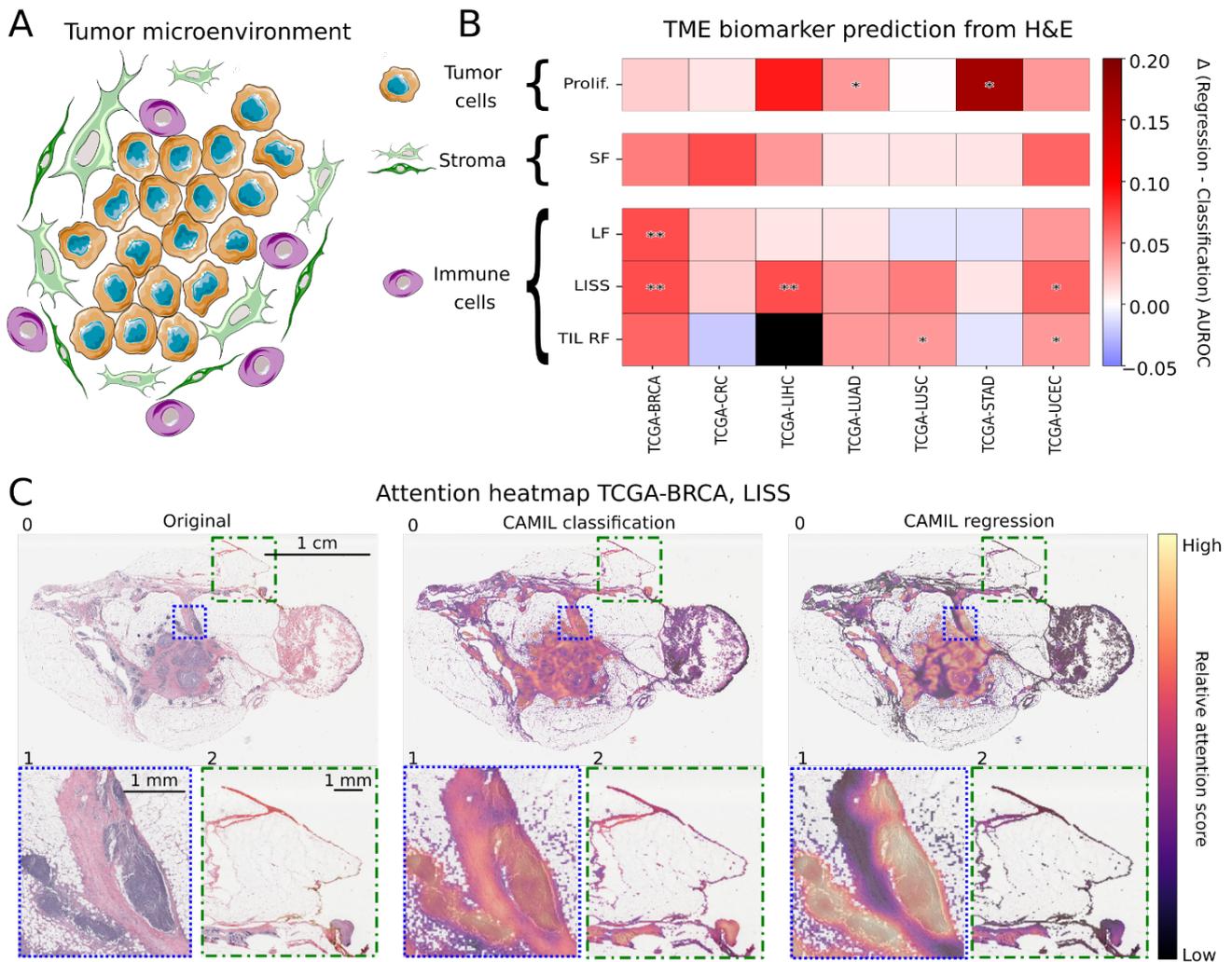

**Figure 3: CAMIL classification versus CAMIL regression for the prediction of continuous biological process biomarkers of the tumor microenvironment.** A) The scope in which we analyzed the tumor microenvironment (TME) consists of tumor cells, stroma and immune cells. B) Heatmap depicting area under the receiver operating curve (AUROC) deltas between CAMIL regression and CAMIL classification for 5 biological process biomarkers (tumor infiltrating lymphocytes regional fraction (TIL RF), proliferation (Prolif.), leukocyte fraction (LF), lymphocytes infiltrating signature score (LISS) and stromal fraction (SF)) on the test sets of breast cancer (BRCA), colorectal cancer (CRC), liver hepatocellular carcinoma (LIHC), lung adenocarcinoma (LUAD), lung squamous cell cancer (LUSC), pancreas adenocarcinoma (PAAD), liver hepatocellular carcinoma (LIHC), stomach cancer (STAD) and endometrial cancer (UCEC) from The Cancer Genome Atlas (TCGA) program for site-aware split folds. The higher the positive delta, the better the CAMIL regression model performed. Statistical significance is indicated with an asterisk as a result of a dependent one-sided t-test (α=0.05). C) Attention heatmap of a slide from the test set of TCGA-BRCA. Image 0 shows the entire slide, with an area of interest for diagnostics in image 1. Image 2 shows an area presumably containing non-essential diagnostics information. This is repeated for the original slide, the attention heatmap using the classification model, and the attention heatmap using our CAMIL regression model in fold 0 for LISS. The higher the attention score of an area, the more important it is for the model's decision making. Icon source: smart.servier.com

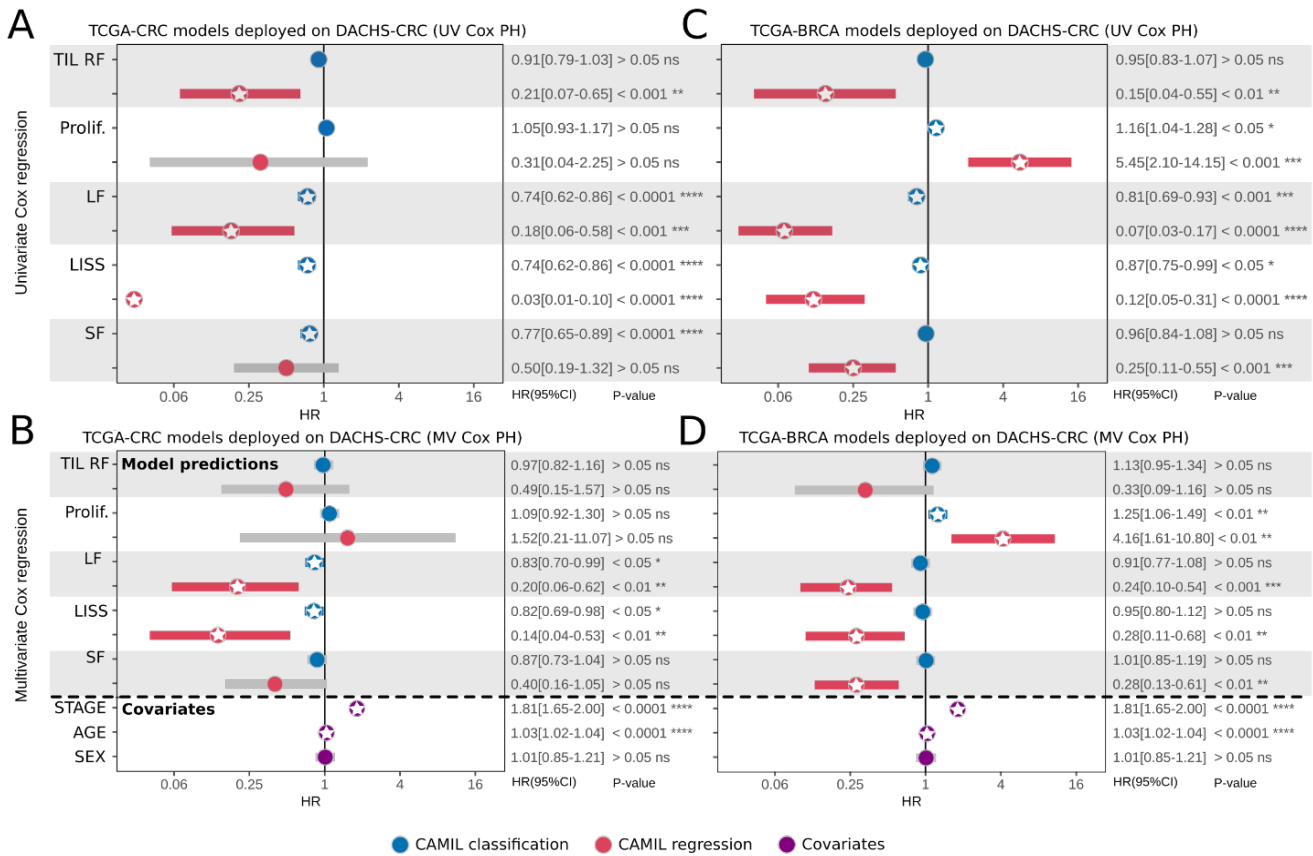

**Figure 4: Overview of the externally validated prognostic capabilities of the trained models to predict overall survival.** Panel A) and B) display an univariate (UV) Cox Proportional-Hazard (PH) analysis of the trained models on The Cancer Genome Atlas (TCGA) program, deployed on the external colorectal cancer (CRC) samples from the Darmkrebs: Chancen der Verhütung durch Screening (DACHS) study for the TCGA-CRC and TCGA breast cancer (BRCA) models, respectively. Panel C) and D) display a multivariate (MV) Cox PH analysis of the trained immune cell models, deployed on the external DACHS-CRC cohort for the TCGA-CRC and TCGA-BRCA models, respectively. Each model's output, from CAMIL classification (categorical class predictions) and CAMIL regression (continuous score predictions), is considered independently together with the three covariates tumor stage, age and sex for the MV Cox PH analysis. The observed biological process biomarkers are tumor infiltrating lymphocytes regional fraction (TIL RF), proliferation (Prolif.), leukocyte fraction (LF), lymphocyte infiltration signature score (LISS), and stromal fraction (SF). Stars indicate statistical significance (p < 0.05) for the hazard ratios (HR) and their 95% confidence interval (CI).

## Author Contributions

OSMEN and JNK designed the study. OSMEN, MVT and MG developed the software. OSMEN, CMLL, TY, MH, HB, AB and JNK contributed to data collection and assembly. OSMEN, ZIC, CMLL, KJH and FK interpreted and analyzed the data. All authors wrote and reviewed the report and approved the final version for submission.

## Funding

JNK is supported by the German Federal Ministry of Health (DEEP LIVER, ZMVI1-2520DAT111) and the Max-Eder-Programme of the German Cancer Aid (grant #70113864), the German Federal Ministry of Education and Research (PEARL, 01KD2104C; CAMINO, 01EO2101; SWAG, 01KD2215A; TRANSFORM LIVER, 031L0312A), the German Academic Exchange Service (SECAI, 57616814), the German Federal Joint Committee (Transplant.KI, 01VSF21048) the European Union (ODELIA, 101057091; GENIAL, 101096312) and the National Institute for Health and Care Research (NIHR, NIHR213331) Leeds Biomedical Research Centre. The views expressed are those of the author(s) and not necessarily those of the NHS, the NIHR or the Department of Health and Social Care.

J.S.R.-F. reports receiving personal/consultancy fees from Goldman Sachs, Bain Capital, REPARE Therapeutics, Saga Diagnostics and Paige.AI, membership of the scientific advisory boards of VolitionRx, REPARE Therapeutics and Paige.AI, membership of the Board of Directors of Grupo Oncoclinicas, and ad hoc membership of the scientific advisory boards of Astrazeneca, Merck, Daiichi Sankyo, Roche Tissue Diagnostics and Personalis, outside the scope of this study.

UCLH Biomedical research centre is funded by the National Institute for Health Research (BRC399/NS/RB/101410). SB is also supported by the Department of Health's NIHR Biomedical Research Centre's funding scheme. TM was supported by The Brain Tumour Charity (GN-000389 clinical research training fellowship) and by the National Institute of health research (NIHR) with clinical lecturer fellowship (CL-2019-19-001).

FRK is supported by the Add-on Fellowship of the Joachim Herz Foundation.

QZ is funded by the China Scholarship Council (Grant n°201908070052).

## Competing Interests

JNK reports consulting services for Owkin, France, Panakeia, UK, and DoMore Diagnostics, Norway and has received honoraria for lectures by MSD, Eisai, and Fresenius.

J.S.R.-F. is funded in part by the Breast Cancer Research Foundation, by a Susan G Komen Leadership grant, and by the NIH/NCI P50 CA247749 01 grant.

# Supplements

## Raw results and statistics

| Cohort | CAMIL classification | | | Regression (Graziani et al.) | | | CAMIL regression | | |
|---|---|---|---|---|---|---|---|---|---|
| | AUROC 95%CI | AUPRC 95%CI | p-value | AUROC 95%CI | AUPRC 95%CI | p-value | AUROC 95%CI | AUPRC 95%CI | p-value |
| **TCGA-BRCA** | 0.74 ± 0.03 | 0.53 ± 0.09 | 2.13E-06 | 0.70 ± 0.03 | 0.46 ± 0.07 | 2.07E-04 | **0.78 ± 0.02** | **0.56 ± 0.05** | **3.08E-09** |
| **TCGA-CRC** | 0.73 ± 0.15 | 0.20 ± 0.21 | 2.44E-01 | 0.47 ± 0.17 | 0.05 ± 0.03 | 5.83E-01 | 0.76 ± 0.12 | 0.16 ± 0.09 | 1.56E-01 |
| **TCGA-GBM** | 0.64 ± 0.49 | 0.18 ± 0.24 | NA (1 sample) | 0.64 ± 0.23 | 0.11 ± 0.16 | NA (1 sample) | 0.75 ± 0.40 | 0.48 ± 0.61 | NA (1 sample) |
| **TCGA-LUAD** | 0.70 ± 0.05 | 0.57 ± 0.12 | 1.39E-02 | 0.71 ± 0.03 | 0.59 ± 0.05 | 2.37E-03 | **0.72 ± 0.05** | **0.57 ± 0.12** | **4.44E-03** |
| **TCGA-LUSC** | 0.57 ± 0.03 | 0.64 ± 0.07 | 3.50E-01 | 0.57 ± 0.04 | 0.61 ± 0.06 | 3.49E-01 | 0.57 ± 0.05 | 0.62 ± 0.07 | 3.12E-01 |
| **TCGA-PAAD** | 0.54 ± 0.32 | 0.20 ± 0.22 | 3.66E-01 | 0.40 ± 0.17 | 0.15 ± 0.17 | 5.14E-01 | 0.72 ± 0.06 | 0.19 ± 0.06 | 3.06E-01 |
| **TCGA-UCEC** | 0.73 ± 0.05 | 0.33 ± 0.11 | 1.08E-02 | 0.67 ± 0.11 | 0.30 ± 0.13 | 1.39E-01 | **0.82 ± 0.03** | **0.39 ± 0.06** | **2.15E-04** |
| **CPTAC-LUAD** | 0.82 ± 0.04 | 0.36 ± 0.07 | 3.63E-04 | **0.83 ± 0.03** | **0.45 ± 0.04** | **5.06E-05** | 0.81 ± 0.03 | 0.43 ± 0.03 | 1.40E-04 |
| **CPTAC-LUSC** | 0.62 ± 0.04 | 0.43 ± 0.03 | 9.14E-02 | **0.66 ± 0.03** | **0.49 ± 0.06** | **1.30E-02** | 0.62 ± 0.06 | 0.45 ± 0.06 | 1.00E-01 |
| **CPTAC-PAAD** | 0.60 ± 0.32 | 0.23 ± 0.33 | 3.21E-01 | 0.55 ± 0.34 | 0.07 ± 0.07 | 2.89E-01 | 0.68 ± 0.04 | 0.09 ± 0.03 | 2.25E-01 |
| **CPTAC-UCEC** | **0.98 ± 0.02** | **0.66 ± 0.17** | **6.50E-04** | 0.89 ± 0.14 | 0.25 ± 0.03 | 4.67E-02 | 0.96 ± 0.01 | 0.50 ± 0.13 | 2.19E-05 |

**Suppl. Table 1: Area under the receiver operating characteristics (AUROC) and area under the precision recall characteristics (AUPRC) with 95% confidence interval (CI) and corresponding p-values of the homologous recombination deficiency (HRD) target with site-aware splits.** The evaluation AUROC and AUPRC for this CAMIL classification, Graziani et al. regression and CAMIL regression is calculated for each model trained on the HRD score. In The Cancer Genome Atlas (TCGA), breast cancer (BRCA), colorectal cancer (CRC), glioblastoma (GBM), lung adenocarcinoma (LUAD), lung squamous cell cancer (LUSC), pancreatic adenocarcinoma (PAAD) and endometrial cancer (UCEC) were used for site-aware training. In the Clinical Proteomic Tumor Analysis Consortium (CPTAC) effort, LUAD, LUSC, PAAD and UCEC were used as external validation cohorts. The p-values are a result of an independent two-sided t-test comparing the means of the positive and negative scores of the models' predictions. Statistical significance is reached at p < 0.05 and indicates a difference in the means between the positive and negative scores. Statistically insignificant AUROCs and AUPRCs for the HRD score prediction are marked with gray.

| Cohort | F value | DoF (1) | DoF (2) | p-value |
|---|---:|---:|---:|---:|
| TCGA-BRCA | 89.870968 | 8 | 8 | 3.00E-06 |
| TCGA-CRC | 86.6317 | 8 | 8 | 4.00E-06 |
| TCGA-GBM | 1.4407 | 2 | 8 | 2.92E-01 |
| TCGA-PAAD | 10.2074 | 2 | 8 | 6.30E-03 |
| TCGA-LUAD | 14.236398 | 2 | 8 | 2.32E-03 |
| TCGA-LUSC | 38.110092 | 2 | 8 | 8.10E-05 |
| TCGA-UCEC | 6.9612 | 2 | 8 | 1.77E-02 |
| CPTAC-UCEC | 3.2108 | 2 | 8 | 9.47E-02 |
| CPTAC-LUAD | 6 | 2 | 8 | 2.56E-02 |
| CPTAC-LUSC | 6.9091 | 2 | 8 | 1.81E-02 |
| CPTAC-PAAD | 1.0242 | 2 | 8 | 4.02E-01 |

**Suppl. Table 2: Repeated measures analysis of variance (ANOVA) of the three modeling approaches for the homologous recombination deficiency (HRD) score with site-aware splits.** The repeated measures ANOVA for the models trained on the HRD score through the CAMIL classification, Graziani's regression and our CAMIL regression approach, resulting in F values with the degrees of freedom (DoF) 1 and 2. Statistical significance is reached at $p < 0.05$ and indicates a difference in the means between the three modeling approaches. Statistically insignificant F values for the models trained on the HRD score are marked with gray, with. In The Cancer Genome Atlas (TCGA), breast cancer (BRCA), colorectal cancer (CRC), glioblastoma (GBM), lung adenocarcinoma (LUAD), lung squamous cell cancer (LUSC), pancreatic adenocarcinoma (PAAD) and endometrial cancer (UCEC) were used for site-aware training. In the Clinical Proteomic Tumor Analysis Consortium (CPTAC) effort, LUAD, LUSC, PAAD and UCEC were used as external validation cohorts

|  | CAMIL regression > CAMIL classification | | Regression (Graziani et al.) > CAMIL classification | |
| --- | --- | --- | --- | --- |
|  | statistic | P-value | statistic | P-value |
| **TCGA-GBM** | 0.9112295657 | 2.07E-01 |  |  |
| **TCGA-PAAD** | 7.212386465 | 9.80E-04 |  |  |
| **TCGA-UCEC** | 5.049515586 | 3.62E-03 |  |  |
| **TCGA-CRC** | 12.14664495 | 1.32E-04 |  |  |
| **TCGA-BRCA** | 4.75 | 4.49E-03 |  |  |
| **TCGA-LUAD** | 2.666666667 | 2.80E-02 |  |  |
| **TCGA-LUSC** | 0.5827715174 | 2.96E-01 |  |  |
| **CPTAC-UCEC** | -6.32455532 | 1.18E-01 | -2.0197547 | 9.43E-01 |
| **CPTAC-LUAD** | -1.870828693 | 9.33E-01 | 1.206045378 | 1.47E-01 |
| **CPTAC-LUSC** | 2.48E-15 | 5.00E-01 | 4.146139914 | 7.15E-03 |
| **CPTAC-PAAD** | 0.7372282207 | 2.51E-01 | -2.411214111 | 9.63E-01 |

**Suppl. Table 3: One-sided dependent t-tests to determine if regression outperforms classification in the models for homologous recombination deficiency (HRD) prediction using site-aware splits.** For predicting HRD, models were trained in a site-aware manner on The Cancer Genome Atlas (TCGA), breast cancer (BRCA), colorectal cancer (CRC), glioblastoma (GBM), lung adenocarcinoma (LUAD), lung squamous cell cancer (LUSC), pancreatic adenocarcinoma (PAAD) and endometrial cancer (UCEC). In the Clinical Proteomic Tumor Analysis Consortium (CPTAC) effort, LUAD, LUSC, PAAD and UCEC were used as external validation cohorts. In TCGA, CAMIL regression model is compared with a dependent one-sided t-test to the CAMIL classification approach, with significance for $p < 0.05$. In CPTAC, two hypotheses are tested through dependent one-sided t-tests: whether CAMIL regression is better than CAMIL classification, and whether Graziani's regression is better than CAMIL classification. Here, significance is reached at $p < 0.025$ when using Bonferroni's correction for multiple hypothesis testing.

|  | CAMIL classification | | | | CAMIL regression | | | | Improvement using regression | Mean improvement using regression |
|---|---|---|---|---|---|---|---|---|---|---|
|  | HRD+ median | HRD+ IQR | HRD- median | HRD- IQR | HRD+ median | HRD+ IQR | HRD- median | HRD- IQR | | |
| **TCGA-BRCA** | 0.64 | 0.32 | 0.43 | 0.28 | 0.53 | 0.32 | 0.26 | 0.27 | 29.5% | |
| **TCGA-CRC** | 0.46 | 0.10 | 0.45 | 0.05 | 0.44 | 0.12 | 0.35 | 0.17 | 862.0% | |
| **TCGA-GBM** | 0.61 | 0.13 | 0.59 | 0.16 | 0.68 | 0.43 | 0.42 | 0.17 | 1157.0% | |
| **TCGA-LUAD** | 0.52 | 0.16 | 0.44 | 0.16 | 0.57 | 0.17 | 0.44 | 0.21 | 44.8% | |
| **TCGA-LUSC** | 0.45 | 0.09 | 0.44 | 0.09 | 0.58 | 0.23 | 0.55 | 0.22 | 260.3% | |
| **TCGA-PAAD** | 0.66 | 0.25 | 0.64 | 0.29 | 0.65 | 0.09 | 0.57 | 0.18 | 238.2% | |
| **TCGA-UCEC** | 0.66 | 0.20 | 0.51 | 0.23 | 0.37 | 0.31 | 0.14 | 0.20 | 50.5% | 377.5% |
| **CPTAC-LUAD** | 0.68 | 0.16 | 0.52 | 0.23 | 0.64 | 0.15 | 0.50 | 0.19 | -14.6% | |
| **CPTAC-LUSC** | 0.58 | 0.12 | 0.55 | 0.15 | 0.46 | 0.16 | 0.41 | 0.16 | 101.2% | |
| **CPTAC-PAAD** | 0.85 | 0.24 | 0.72 | 0.26 | 0.41 | 0.17 | 0.37 | 0.14 | -64.6% | |
| **CPTAC-UCEC** | 0.82 | 0.13 | 0.44 | 0.26 | 0.72 | 0.30 | 0.14 | 0.19 | 55.5% | 19.4% |

**Suppl. Table 4: Median and interquartile range (IQR) of the min-max normalized predicted scores for CAMIL classification and CAMIL regression approach.** The median and IQR of the min-max normalized prediction scores AUROC for the CAMIL classification and CAMIL regression is calculated for each model trained on the HRD score. A positive percentage indicates a larger distance between the median peaks of the groups' distribution using regression, whereas a negative percentage indicates a larger distance between the median peaks of the groups' distribution using classification. In The Cancer Genome Atlas (TCGA), breast cancer (BRCA), colorectal cancer (CRC), glioblastoma (GBM), lung adenocarcinoma (LUAD), lung squamous cell cancer (LUSC), pancreatic adenocarcinoma (PAAD) and endometrial cancer (UCEC) were used for site-aware training. In the Clinical Proteomic Tumor Analysis Consortium (CPTAC) effort, LUAD, LUSC, PAAD and UCEC were used as external validation cohorts.

|  | Regression (Graziani et al.) | | CAMIL regression | |
| --- | --- | --- | --- | --- |
|  | $R^2$ | p-value | $R^2$ | p-value |
| **TCGA-BRCA** | 0.001802193733 | 1.94E-01 | **0.2002149262** | **2.02E-47** |
| **TCGA-CRC** | 0.01161051257 | 1.95E-02 | **0.05613856911** | **2.02E-07** |
| **TCGA-GBM** | 0.0008700403322 | 6.61E-01 | **0.03640186757** | **4.24E-03** |
| **TCGA-LUAD** | 0.09925245916 | 6.78E-12 | **0.2161578859** | **1.13E-25** |
| **TCGA-LUSC** | 0.0007656344841 | 5.72E-01 | **0.02017953956** | **3.57E-03** |
| **TCGA-PAAD** | 9.89E-05 | 8.97E-01 | **0.03608239068** | **1.26E-02** |
| **TCGA-UCEC** | 7.89E-05 | 8.47E-01 | **0.1361695292** | **9.06E-17** |
| **CPTAC-LUSC** | 1.96E-06 | 9.74E-01 | **0.01438696255** | **5.26E-03** |
| **CPTAC-LUAD** | 0.004616549633 | 1.18E-01 | **0.1310443931** | **7.44E-18** |
| **CPTAC-PAAD** | 1.94E-05 | 9.08E-01 | **0.01012561025** | **7.94E-03** |
| **CPTAC-UCEC** | 3.37E-05 | 8.97E-01 | **0.2625652048** | **1.73E-34** |

**Suppl. Table 5: Comparison of Graziani regression with CAMIL regression through the coefficient of determination ($R^2$).** The p-value follows from a two-sided independent t-test with significance reached at $p < 0.05$, indicating the correlation coefficient is non-zero. Within a range of 0 to 1, the higher the $R^2$, the better the regression model. In The Cancer Genome Atlas (TCGA), breast cancer (BRCA), colorectal cancer (CRC), glioblastoma (GBM), lung adenocarcinoma (LUAD), lung squamous cell cancer (LUSC), pancreatic adenocarcinoma (PAAD) and endometrial cancer (UCEC) were used for site-aware training. In the Clinical Proteomic Tumor Analysis Consortium (CPTAC) effort, LUAD, LUSC, PAAD and UCEC were used as external validation cohorts.

## CAMIL regression has better generalization capabilities than Graziani regression

Given the nature of how AUROCs are produced, the continuous output score of the regression models can be used in combination with a categorical target, such as the clinically-relevant binarized HRD score. However, AUROCs only indicate how well the given continuous score is able to separate between the negative and positive class, i.e. rewarding a high AUROC for a model which outputs a low intra-class variance and a high inter-class variance, regardless of the absolute range of the prediction scores. Analyzing the performance between regression methods, it was found that CAMIL regression is capable of predicting more clinically-relevant output scores which are closer to the absolute ground-truth (**Suppl. Fig. 1)**, giving a prediction range of (29, 34) and (10, 55) for the Graziani et al. regression model and our CAMIL regression model on the CPTAC-LUAD external test cohort, respectively. For this analysis, LUAD was chosen as it resulted in the only statistically significant tumor-type for the regression models with both an internal and external validation set for the HRD target. With an $R^2$ of 0.16 (3,06E-22) for the Graziani et al. regression and an $R^2$ of 0.29 (1.39E-40) for our proposed CAMIL regression model, superior generalization capabilities for our proposed CAMIL regression over the regression method by Graziani et al. are observed. However, the measure by AUROC would indicate that the regression by Graziani et al. has superior performance, showing the limited capabilities of comparing regression models solely through AUROC values.

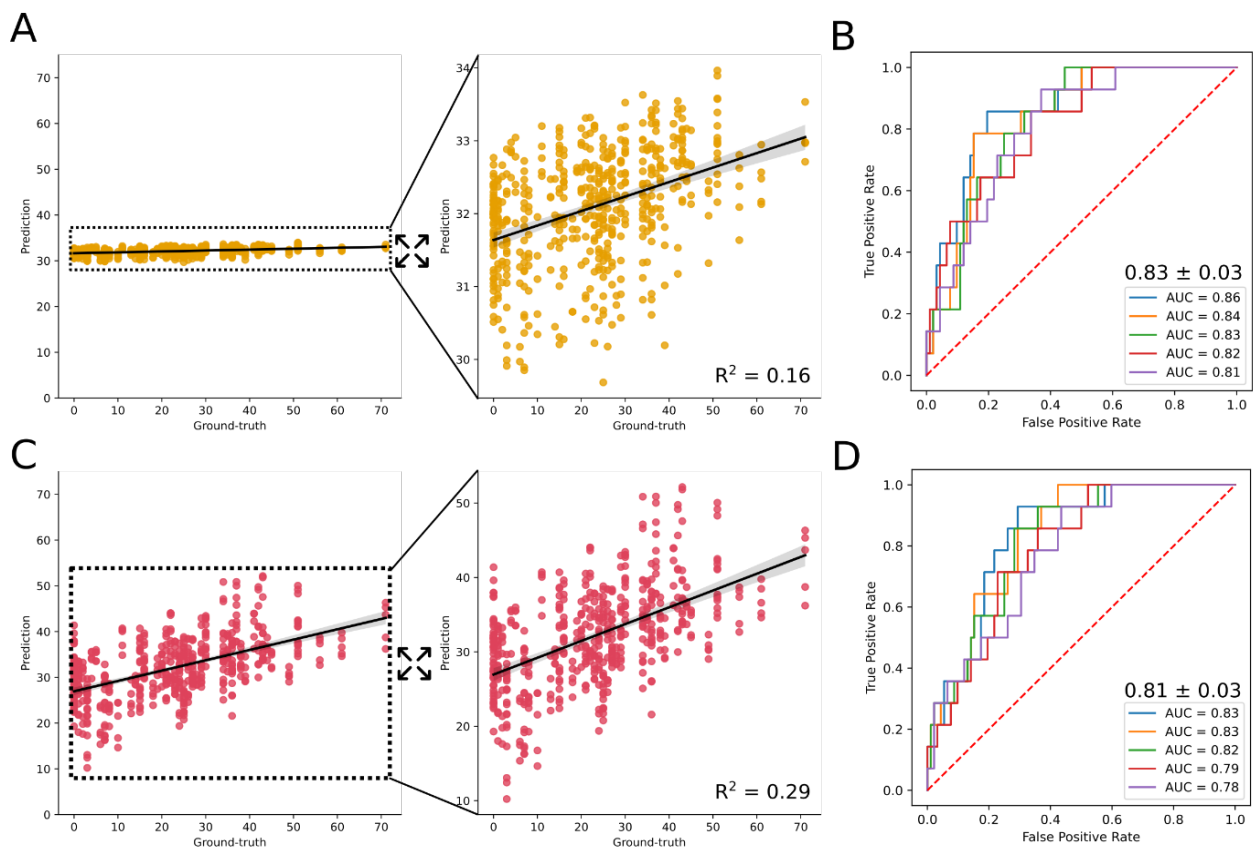

**Suppl. Figure 1: Comparison of CAMIL regression and Graziani et al. regression on an external testing cohort.** A) The correlation plot of the regression approach by Graziani et al. on the external cohort of lung adenocarcinoma (LUAD) from the Clinical Proteomic Tumor Analysis Consortium (CPTAC) effort in the original range of the homologous recombination deficiency (HRD) continuous ground-truth, and a zoom-in of the same data in the range of the model's prediction. B) The corresponding area under the receiver operator characteristic (AUROC) curve using the continuous prediction scores plotted in panel A and the HRD binary ground-truth. C) The correlation plot of the CAMIL regression approach from this study on the external cohort CPTAC-LUAD in the original range of the HRD continuous ground-truth, and a zoom-in of the same data in the range of the model's prediction. D) The corresponding AUROC curve using the continuous prediction scores plotted in panel C and the HRD binary ground-truth.

|  | CAMIL classification | Graziani et al. regression | CAMIL classification |
|---|---|---|---|
| batch size | 64 | 1 | 1 |
| batch normalization | Yes | No | No |
| optimizer | Adam | Stochastic gradient descent | Adam |
| loss function | Weighted Cross-entropy | Mean Squared Error | Weighted Mean Squared Error |
| epochs | 25 | 100 | 25 |
| dropout | 50% | 20% | 0% |
| target balancing | Inverse weighted | No | Kernel-based[42] |
| Base model | ImageNet weighted ResNet50 + contrastive clustering[26] | ImageNet weighted ResNet18 | ImageNet weighted ResNet50 + contrastive clustering[26] |

**Suppl. Table 6: Overview of the differences between the three modeling approaches.** The main differences between contrastively-clustered attention-based multiple instance learning (CAMIL) classification, Graziani et al. regression and CAMIL regression.

For deeper analysis into the differences between the regression heads of Graziani et al. regression and CAMIL regression, an ablation study (**Suppl. Table 7)** was performed on CAMIL regression using the TCGA-BRCA cohort. The TCGA-BRCA cohort was chosen for the ablation study as both Graziani et al. regression model and the CAMIL regression model gave statistically significant AUROCs for all 5 folds which were in a similar range with low variance, in contrast to TCGA-LUAD which showed more variance among the 5 folds due to an outlying fold (**Fig. 2**).

|  | [0,91] prediction range | [0,1] $R^2$ mean ± std | [0,1] AUROC 95% CI |
|---|---|---|---|
| CAMIL regression | (6.68, 69.71) | 0.30 ± 0.02 (p-value 2.65e-72) | 0.78 ± 0.02 (p-value 3.08e-09) |
| With 20% dropout | (7.03, 70.09) | 0.30 ± 0.03 (p-value 2.90e-72) | 0.78 ± 0.02 (p-value 1.83e-09) |
| With SGD | ***(33.71, 37.12)*** | ***0.04 ± 0.02 (p-value 1.21e-01)*** | ***0.62 ± 0.07 (p-value 0.08)*** |
| With 100 epochs | (6.08, 67.89) | 0.30 ± 0.02 (p-value 3.43e-73) | 0.78 ± 0.01 (p-value 3.00e-09) |
| Without kernel-based balancing | (6.91, 68.93) | 0.30 ± 0.02 (p-value 4.59e-71) | 0.78 ± 0.02 (p-value 4.42e-09) |

**Suppl. Table 7: Ablation study of CAMIL regression with adaptations from Graziani et al. regression for the test set of The Cancer Genome Atlas (TCGA) breast cancer cohort.** Using the same site-aware splits for all the models, our CAMIL regression modeling approach was altered according to the changes found in Graziani et al. regression approach, adding a layer with 20% dropout, swapping the Adam optimizer for stochastic gradient descent (SGD), increasing the epochs to 100, and removing the kernel-based balancing. The regression approaches are compared through their prediction score range between [0, 91], where a wider range is better, the coefficient of determination ($R^2$) with standard deviation (std), and the area under the receiver operating characteristic (AUROC) with 95% confidence interval (CI) across all five splits.

**CAMIL regression AUROC**

| | TCGA-BRCA | TCGA-CRC | TCGA-LIHC | TCGA-LUAD | TCGA-LUSC | TCGA-STAD | TCGA-UCEC |
|---|---|---|---|---|---|---|---|
| **TIL Regional Fraction** | 0.88 ± 0.02 | 0.80 ± 0.05 | NA | 0.89 ± 0.06 | 0.91 ± 0.01 | 0.86 ± 0.03 | 0.85 ± 0.05 |
| **Proliferation** | 0.83 ± 0.05 | 0.63 ± 0.08 | 0.85 ± 0.09 | 0.81 ± 0.06 | 0.73 ± 0.19 | 0.90 ± 0.05 | 0.74 ± 0.08 |
| **Leukocyte Fraction** | 0.81 ± 0.03 | 0.75 ± 0.05 | 0.80 ± 0.08 | 0.76 ± 0.10 | 0.74 ± 0.07 | 0.72 ± 0.04 | 0.76 ± 0.04 |
| **LISS** | 0.80 ± 0.03 | 0.70 ± 0.07 | 0.73 ± 0.07 | 0.75 ± 0.07 | 0.74 ± 0.07 | 0.70 ± 0.11 | 0.73 ± 0.03 |
| **Stromal Fraction** | 0.80 ± 0.03 | 0.70 ± 0.08 | 0.81 ± 0.09 | 0.76 ± 0.07 | 0.79 ± 0.10 | 0.65 ± 0.10 | 0.69 ± 0.03 |

**CAMIL regression p-val**

| | TCGA-BRCA | TCGA-CRC | TCGA-LIHC | TCGA-LUAD | TCGA-LUSC | TCGA-STAD | TCGA-UCEC |
|---|---|---|---|---|---|---|---|
| **TIL Regional Fraction** | 1.13E-18 | 3.11E-05 | NA | 1.22E-07 | 2.91E-06 | 8.64E-04 | 2.51E-07 |
| **Proliferation** | 3.54E-16 | 1.79E-01 | 1.36E-03 | 4.45E-05 | 1.46E-01 | 2.35E-06 | 6.54E-03 |
| **Leukocyte Fraction** | 2.66E-13 | 4.14E-04 | 4.02E-04 | 1.93E-02 | 3.22E-03 | 9.47E-03 | 1.79E-04 |
| **LISS** | 1.91E-12 | 6.24E-03 | 4.32E-03 | 3.20E-02 | 4.47E-03 | 7.78E-02 | 8.11E-05 |
| **Stromal Fraction** | 2.53E-11 | 3.08E-02 | 1.47E-03 | 3.08E-03 | 3.64E-03 | 1.67E-01 | 5.83E-03 |

**CAMIL classification AUROC**

| | TCGA-BRCA | TCGA-CRC | TCGA-LIHC | TCGA-LUAD | TCGA-LUSC | TCGA-STAD | TCGA-UCEC |
|---|---|---|---|---|---|---|---|
| **TIL Regional Fraction** | 0.82 ± 0.08 | 0.81 ± 0.05 | NA | 0.83 ± 0.10 | 0.84 ± 0.04 | 0.80 ± 0.14 | 0.78 ± 0.03 |
| **Proliferation** | 0.82 ± 0.05 | 0.58 ± 0.13 | 0.78 ± 0.07 | 0.80 ± 0.06 | 0.78 ± 0.11 | 0.70 ± 0.20 | 0.69 ± 0.08 |
| **Leukocyte Fraction** | 0.73 ± 0.05 | 0.74 ± 0.07 | 0.79 ± 0.04 | 0.71 ± 0.02 | 0.73 ± 0.14 | 0.73 ± 0.08 | 0.73 ± 0.08 |
| **LISS** | 0.73 ± 0.05 | 0.68 ± 0.10 | 0.63 ± 0.03 | 0.61 ± 0.15 | 0.69 ± 0.14 | 0.66 ± 0.06 | 0.67 ± 0.05 |
| **Stromal Fraction** | 0.76 ± 0.02 | 0.61 ± 0.07 | 0.73 ± 0.06 | 0.75 ± 0.05 | 0.76 ± 0.06 | 0.61 ± 0.08 | 0.61 ± 0.07 |

**CAMIL classification p-val**

| | TCGA-BRCA | TCGA-CRC | TCGA-LIHC | TCGA-LUAD | TCGA-LUSC | TCGA-STAD | TCGA-UCEC |
|---|---|---|---|---|---|---|---|
| **TIL Regional Fraction** | 1.29E-16 | 4.57E-07 | NA | 7.24E-10 | 8.62E-07 | 7.57E-04 | 3.09E-07 |
| **Proliferation** | 3.93E-12 | 3.92E-01 | 4.04E-02 | 8.71E-04 | 1.08E-02 | 6.46E-04 | 2.43E-02 |
| **Leukocyte Fraction** | 6.24E-10 | 5.55E-05 | 4.55E-04 | 3.40E-03 | 2.64E-03 | 3.75E-02 | 3.90E-04 |
| **LISS** | 4.03E-09 | 2.60E-02 | 2.12E-02 | 6.39E-02 | 6.62E-02 | 1.61E-01 | 6.58E-03 |
| **Stromal Fraction** | 4.92E-09 | 4.78E-03 | 4.51E-04 | 2.30E-02 | 3.40E-03 | 1.75E-01 | 5.77E-02 |

**Delta AUROC CAMIL regression – CAMIL classification**

| | TCGA-BRCA | TCGA-CRC | TCGA-LIHC | TCGA-LUAD | TCGA-LUSC | TCGA-STAD | TCGA-UCEC |
|---|---|---|---|---|---|---|---|
| **TIL Regional Fraction** | 0.06 | -0.01 | NA | 0.06 | 0.07 | 0.06 | 0.07 |
| **Proliferation** | 0.01 | 0.05 | 0.07 | 0.01 | -0.05 | 0.2 | 0.05 |
| **Leukocyte Fraction** | 0.08 | 0.01 | 0.01 | 0.05 | 0.01 | -0.01 | 0.03 |
| **Lymphocyte Infiltration Signature** | 0.07 | 0.02 | 0.1 | 0.14 | 0.05 | 0.04 | 0.06 |

| Score | | | | | | | |
|---|---|---|---|---|---|---|---|
| Stromal Fraction | 0.04 | 0.09 | 0.08 | 0.01 | 0.03 | 0.04 | 0.08 |

**Suppl. Table 8: Area under the receiver operating curve (AUROC) and p-values with 95% confidence interval with patient-level splits.** The performance of the CAMIL classification and CAMIL regression models without site-aware splits is measured on cohorts from The Cancer Genome Atlas (TCGA), breast cancer (BRCA), colorectal cancer (CRC), liver hepatocellular carcinoma (LIHC), lung adenocarcinoma (LUAD), lung squamous cell cancer (LUSC), gastric cancer (STAD) and endometrial cancer (UCEC) on biomarkers for tumor infiltrating lymphocytes (TIL) regional fraction, proliferation, leukocyte fraction, lymphocyte infiltration signature score, and stromal fraction. The performance metric is the AUROC with corresponding 95%CI and p-values.

**CAMIL regression AUROC**

| | TCGA-BRCA | TCGA-CRC | TCGA-LIHC | TCGA-LUAD | TCGA-LUSC | TCGA-STAD | TCGA-UCEC |
|---|---|---|---|---|---|---|---|
| **TIL Regional Fraction** | 0.88 ± 0.02 | 0.80 ± 0.05 | NA | 0.89 ± 0.06 | 0.91 ± 0.01 | 0.86 ± 0.03 | 0.85 ± 0.05 |
| **Proliferation** | 0.83 ± 0.05 | 0.63 ± 0.08 | 0.85 ± 0.09 | 0.81 ± 0.06 | 0.73 ± 0.19 | 0.90 ± 0.05 | 0.74 ± 0.08 |
| **Leukocyte Fraction** | 0.81 ± 0.03 | 0.75 ± 0.05 | 0.80 ± 0.08 | 0.76 ± 0.10 | 0.74 ± 0.07 | 0.72 ± 0.04 | 0.76 ± 0.04 |
| **LISS** | 0.80 ± 0.03 | 0.70 ± 0.07 | 0.73 ± 0.07 | 0.75 ± 0.07 | 0.74 ± 0.07 | 0.70 ± 0.11 | 0.73 ± 0.03 |
| **Stromal Fraction** | 0.80 ± 0.03 | 0.70 ± 0.08 | 0.81 ± 0.09 | 0.76 ± 0.07 | 0.79 ± 0.10 | 0.65 ± 0.10 | 0.69 ± 0.03 |

**CAMIL regression p-val**

| | TCGA-BRCA | TCGA-CRC | TCGA-LIHC | TCGA-LUAD | TCGA-LUSC | TCGA-STAD | TCGA-UCEC |
|---|---|---|---|---|---|---|---|
| **TIL Regional Fraction** | 1.13E-18 | 3.11E-05 | NA | 1.22E-07 | 2.91E-06 | 8.64E-04 | 2.51E-07 |
| **Proliferation** | 3.54E-16 | 1.79E-01 | 1.36E-03 | 4.45E-05 | 1.46E-01 | 2.35E-06 | 6.54E-03 |
| **Leukocyte Fraction** | 2.66E-13 | 4.14E-04 | 4.02E-04 | 1.93E-02 | 3.22E-03 | 9.47E-03 | 1.79E-04 |
| **LISS** | 1.91E-12 | 6.24E-03 | 4.32E-03 | 3.20E-02 | 4.47E-03 | 7.78E-02 | 8.11E-05 |
| **Stromal Fraction** | 2.53E-11 | 3.08E-02 | 1.47E-03 | 3.08E-03 | 3.64E-03 | 1.67E-01 | 5.83E-03 |

**CAMIL classification AUROC**

| | TCGA-BRCA | TCGA-CRC | TCGA-LIHC | TCGA-LUAD | TCGA-LUSC | TCGA-STAD | TCGA-UCEC |
|---|---|---|---|---|---|---|---|
| **TIL Regional Fraction** | 0.82 ± 0.08 | 0.81 ± 0.05 | NA | 0.83 ± 0.10 | 0.84 ± 0.04 | 0.80 ± 0.14 | 0.78 ± 0.03 |
| **Proliferation** | 0.82 ± 0.05 | 0.58 ± 0.13 | 0.78 ± 0.07 | 0.80 ± 0.06 | 0.78 ± 0.11 | 0.70 ± 0.20 | 0.69 ± 0.08 |
| **Leukocyte Fraction** | 0.73 ± 0.05 | 0.74 ± 0.07 | 0.79 ± 0.04 | 0.71 ± 0.02 | 0.73 ± 0.14 | 0.73 ± 0.08 | 0.73 ± 0.08 |
| **LISS** | 0.73 ± 0.05 | 0.68 ± 0.10 | 0.63 ± 0.03 | 0.61 ± 0.15 | 0.69 ± 0.14 | 0.66 ± 0.06 | 0.67 ± 0.05 |
| **Stromal Fraction** | 0.76 ± 0.02 | 0.61 ± 0.07 | 0.73 ± 0.06 | 0.75 ± 0.05 | 0.76 ± 0.06 | 0.61 ± 0.08 | 0.61 ± 0.07 |

**CAMIL classification p-val**

| | TCGA-BRCA | TCGA-CRC | TCGA-LIHC | TCGA-LUAD | TCGA-LUSC | TCGA-STAD | TCGA-UCEC |
|---|---|---|---|---|---|---|---|
| **TIL Regional Fraction** | 1.29E-16 | 4.57E-07 | NA | 7.24E-10 | 8.62E-07 | 7.57E-04 | 3.09E-07 |
| **Proliferation** | 3.93E-12 | 3.92E-01 | 4.04E-02 | 8.71E-04 | 1.08E-02 | 6.46E-04 | 2.43E-02 |
| **Leukocyte Fraction** | 6.24E-10 | 5.55E-05 | 4.55E-04 | 3.40E-03 | 2.64E-03 | 3.75E-02 | 3.90E-04 |
| **LISS** | 4.03E-09 | 2.60E-02 | 2.12E-02 | 6.39E-02 | 6.62E-02 | 1.61E-01 | 6.58E-03 |
| **Stromal Fraction** | 4.92E-09 | 4.78E-03 | 4.51E-04 | 2.30E-02 | 3.40E-03 | 1.75E-01 | 5.77E-02 |

**Delta AUROC CAMIL regression – CAMIL classification**

| | TCGA-BRCA | TCGA-CRC | TCGA-LIHC | TCGA-LUAD | TCGA-LUSC | TCGA-STAD | TCGA-UCEC |
|---|---|---|---|---|---|---|---|
| **TIL Regional Fraction** | 0.06 | -0.01 | NA | 0.06 | 0.07 | 0.06 | 0.07 |
| **Proliferation** | 0.01 | 0.05 | 0.07 | 0.01 | -0.05 | 0.2 | 0.05 |
| **Leukocyte Fraction** | 0.08 | 0.01 | 0.01 | 0.05 | 0.01 | -0.01 | 0.03 |
| **Lymphocyte Infiltration Signature** | 0.07 | 0.02 | 0.1 | 0.14 | 0.05 | 0.04 | 0.06 |

| Score | | | | | | | |
|---|---|---|---|---|---|---|---|
| Stromal Fraction | 0.04 | 0.09 | 0.08 | 0.01 | 0.03 | 0.04 | 0.08 |

**Suppl. Table 9: Area under the receiver operating curve (AUROC) and p-values with 95% confidence interval (CI) with site-aware splits.** The performance of CAMIL classification and CAMIL regression models with site-aware splits is measured on cohorts from The Cancer Genome Atlas (TCGA), breast cancer (BRCA), colorectal cancer (CRC), liver hepatocellular carcinoma (LIHC), lung adenocarcinoma (LUAD), lung squamous cell cancer (LUSC), gastric cancer (STAD) and endometrial cancer (UCEC) on biomarkers for tumor infiltrating lymphocytes (TIL) regional fraction, proliferation, leukocyte fraction, lymphocyte infiltration signature score, and stromal fraction. The performance metric is the AUROC with corresponding 95%CI and p-values.

| Cohort | LISS | | LF | | Prolif. | | SF | | TIL RF | |
|---|---|---|---|---|---|---|---|---|---|---|
| | T-statistic | P-value | T-statistic | P-value | T-statistic | P-value | T-statistic | P-value | T-statistic | P-value |
| **TCGA-BRCA** | 4.93E+00 | **7.90E-03** | 7.07E+00 | **2.11E-03** | 6.14E+00 | **3.57E-03** | 3.40E+00 | **2.73E-02** | 4.28E+00 | **1.28E-02** |
| **TCGA-CRC** | 2.34E+00 | 7.91E-02 | 2.94E-01 | 7.84E-01 | 1.63E+00 | 1.78E-01 | -1.10E+00 | 3.32E-01 | -1.71E-02 | 9.87E-01 |
| **TCGA-LIHC** | 1.51E+00 | 2.06E-01 | 8.59E-01 | 4.39E-01 | 3.32E+00 | **2.95E-02** | 5.78E-01 | 5.94E-01 | NA | NA |
| **TCGA-LUAD** | 1.89E+00 | 1.31E-01 | 8.93E-01 | 4.22E-01 | 4.09E+00 | **1.50E-02** | 8.95E-01 | 4.21E-01 | 1.61E+00 | 1.82E-01 |
| **TCGA-LUSC** | 3.07E+00 | **3.72E-02** | 1.59E+00 | 1.88E-01 | -7.94E-01 | 4.72E-01 | 1.08E+00 | 3.43E-01 | 1.09E+01 | **4.05E-04** |
| **TCGA-STAD** | 4.66E-01 | 6.66E-01 | 3.59E-01 | 7.38E-01 | 3.19E+00 | **3.33E-02** | -2.81E-01 | 7.93E-01 | -1.02E-01 | 9.23E-01 |
| **TCGA-UCEC** | 2.78E+00 | **4.99E-02** | 3.10E+00 | **3.62E-02** | 3.31E+00 | **2.97E-02** | 1.13E+00 | 3.23E-01 | 1.75E+00 | 1.54E-01 |

**Suppl. Table 10: Dependent two-sided t-test with 95% confidence interval with patient-level splits.** For the models which were trained without site-aware splits, a two-sided dependent t-test was performed to determine whether the means of the area under the receiver operating characteristics (AUROC) from the CAMIL classification and CAMIL regression approach were significantly different (p < 0.05). The models were trained on cohorts from The Cancer Genome Atlas (TCGA), breast cancer (BRCA), colorectal cancer (CRC), liver hepatocellular carcinoma (LIHC), lung adenocarcinoma (LUAD), lung squamous cell cancer (LUSC), gastric cancer (STAD) and endometrial cancer (UCEC) on biomarkers for tumor infiltrating lymphocytes regional fraction (TIL RF), proliferation, leukocyte fraction (LF), lymphocyte infiltration signature score (LISS), and stromal fraction (SF).

| Cohort | LISS | | LF | | Prolif. | | SF | | TIL RF | |
|---|---|---|---|---|---|---|---|---|---|---|
| | T-statistic | P-value | T-statistic | P-value | T-statistic | P-value | T-statistic | P-value | T-statistic | P-value |
| **TCGA-BRCA** | 5.02E+00 | **7.37E-03** | 5.69E+00 | **4.71E-03** | 1.35E+00 | 2.50E-01 | 3.36E+00 | **2.82E-02** | 2.49E+00 | 6.74E-02 |
| **TCGA-CRC** | 1.36E+00 | 2.46E-01 | 7.97E-01 | 4.70E-01 | 6.51E-01 | 5.50E-01 | 2.00E+00 | 1.16E-01 | -1.26E+00 | 2.75E-01 |
| **TCGA-LIHC** | 7.31E+00 | **1.86E-03** | 5.14E-01 | 6.34E-01 | 2.73E+00 | 5.23E-02 | 3.85E+00 | **1.82E-02** | NA | NA |
| **TCGA-LUAD** | 1.17E+00 | 3.07E-01 | 7.16E-01 | 5.13E-01 | 2.84E+00 | **4.67E-02** | 7.15E-01 | 5.14E-01 | 2.33E+00 | 8.05E-02 |
| **TCGA-LUSC** | 1.77E+00 | 1.51E-01 | -2.45E-01 | 8.19E-01 | 2.02E-03 | 9.98E-01 | 3.38E-01 | 7.53E-01 | 3.08E+00 | **3.69E-02** |
| **TCGA-STAD** | 4.42E-01 | 6.82E-01 | -3.87E-01 | 7.19E-01 | 3.44E+00 | **2.63E-02** | 5.15E-01 | 6.34E-01 | -2.03E-01 | 8.49E-01 |
| **TCGA-UCEC** | 4.23E+00 | **1.34E-02** | 1.67E+00 | 1.70E-01 | 1.20E+00 | 2.98E-01 | 1.92E+00 | 1.27E-01 | 3.06E+00 | **3.77E-02** |

**Suppl. Table 11: Dependent two-sided t-test with 95% confidence interval for CAMIL classification and CAMIL regression models trained with site-aware splits.** For the models which were trained without site-aware splits, a two-sided dependent t-test was performed to determine whether the means of the area under the receiver operating characteristics (AUROC) from the CAMIL classification and CAMIL regression approach were significantly different (p < 0.05). The models were trained on cohorts from The Cancer Genome Atlas (TCGA), breast cancer (BRCA), colorectal cancer (CRC), liver hepatocellular carcinoma (LIHC), lung adenocarcinoma (LUAD), lung squamous cell cancer (LUSC), gastric cancer (STAD) and endometrial cancer (UCEC) on biological process biomarkers: tumor infiltrating lymphocytes regional fraction (TIL RF), proliferation (Prolif.), leukocyte fraction (LF), lymphocyte infiltration signature score (LISS), and stromal fraction (SF).

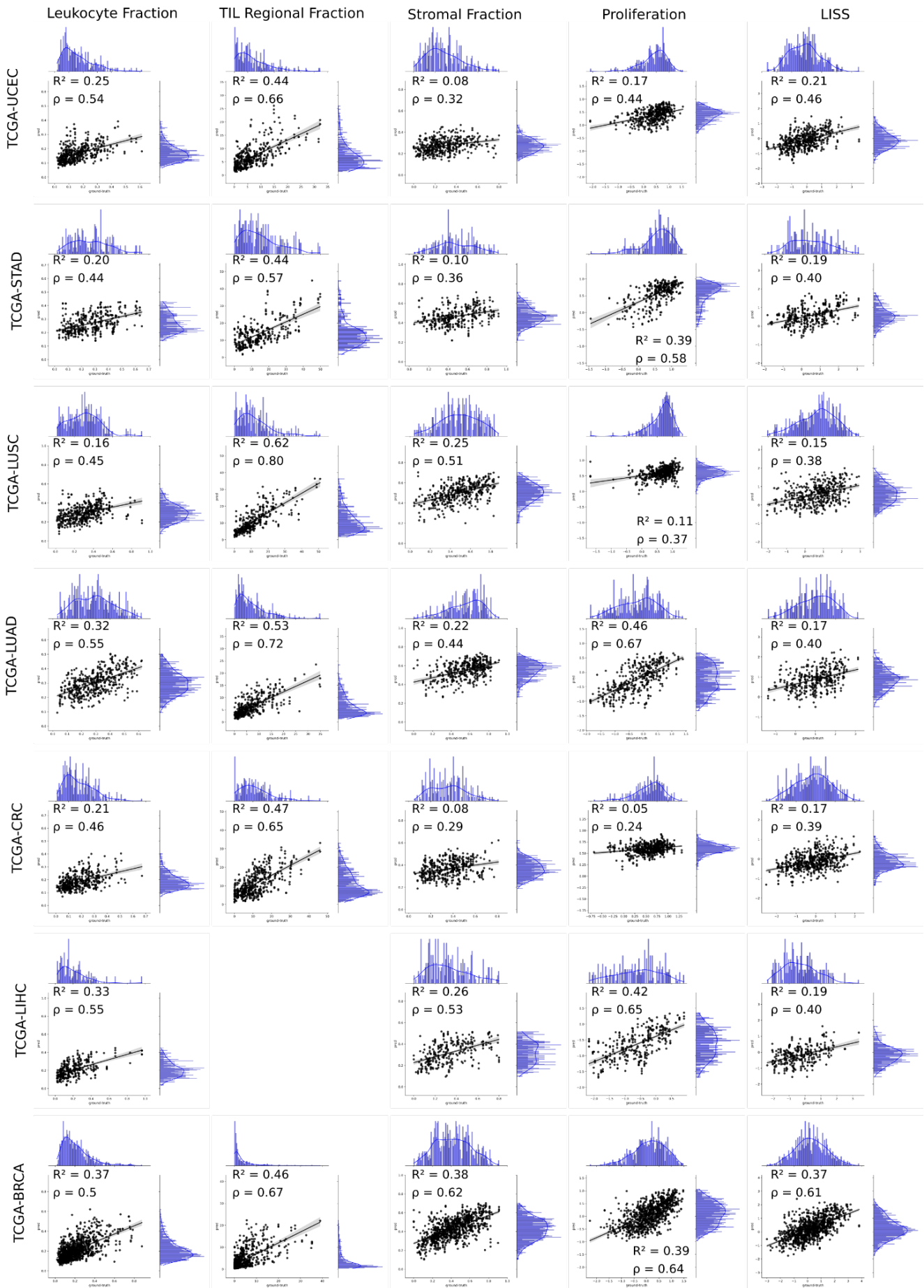

**Suppl. Figure 2: CAMIL regression models' distribution plots with the coefficient of determination ($R^2$) and Spearman's rank correlation coefficient (ρ).** The CAMIL regression

models were trained using site-aware splits on cohorts from The Cancer Genome Atlas (TCGA), breast cancer (BRCA), colorectal cancer (CRC), liver hepatocellular carcinoma (LIHC), lung adenocarcinoma (LUAD), lung squamous cell cancer (LUSC), gastric cancer (STAD) and endometrial cancer (UCEC) on biomarkers for tumor infiltrating lymphocytes (TIL RF) regional fraction, leukocyte fraction, lymphocyte infiltration signature score (LISS), and stromal fraction. The $R^2$ and ρ are shown as performance metrics for the test set of all five site-aware splits. The model's score distributions are displayed on the y-axis, whereas the ground-truth continuous score is displayed on the x-axis.

|  | TCGA-CRC on DACHS-CRC | | | | TCGA-BRCA on DACHS-CRC | | | |
|---|---|---|---|---|---|---|---|---|
| CAMIL model | HR | 95%CI Low | 95%CI High | p-value | HR | 95%CI Low | 95%CI High | p-value |
| **Classification – SF** | 0.77 | 0.65 | 0.89 | **2.14E-05** | 0.96 | 0.84 | 1.08 | 5.40E-01 |
| **Regression – SF** | 0.50 | 0.19 | 1.32 | 1.61E-01 | 0.25 | 0.11 | 0.55 | **5.86E-04** |
| **Classification – LISS** | 0.74 | 0.62 | 0.86 | **1.35E-06** | 0.87 | 0.75 | 0.99 | **2.23E-02** |
| **Regression – LISS** | 0.03 | 0.01 | 0.10 | **9.89E-08** | 0.12 | 0.05 | 0.31 | **8.30E-06** |
| **Classification – LF** | 0.74 | 0.62 | 0.86 | **1.32E-06** | 0.81 | 0.69 | 0.93 | **5.81E-04** |
| **Regression – LF** | 0.18 | 0.06 | 0.58 | **4.13E-03** | 0.07 | 0.03 | 0.17 | **1.14E-09** |
| **Classification – Prolif.** | 1.05 | 0.93 | 1.17 | 3.90E-01 | 1.16 | 1.04 | 1.28 | **1.70E-02** |
| **Regression – Prolif.** | 0.31 | 0.04 | 2.25 | 2.46E-01 | 5.45 | 2.10 | 14.15 | **4.98E-04** |
| **Classification – TIL RF** | 0.91 | 0.79 | 1.03 | 1.20E-01 | 0.95 | 0.83 | 1.07 | 3.90E-01 |
| **Regression – TIL RF** | 0.21 | 0.07 | 0.65 | **7.15E-03** | 0.15 | 0.04 | 0.55 | **4.17E-03** |

**Suppl. Table 12: Hazard ratios (HR) with 95% confidence interval (CI) and corresponding p-values of univariate Cox proportional-hazard models for the models for biological process biomarkers with continuous scores for regression.** The CAMIL regression and CAMIL classification models were trained using site-aware splits on cohorts from The Cancer Genome Atlas colorectal cancer (CRC) and breast cancer (BRCA) on biomarkers for tumor infiltrating lymphocytes regional fraction (TIL RF), proliferation (Prolif.), leukocyte fraction (LF), lymphocyte infiltration signature score (LISS), and stromal fraction (SF). These models were deployed on colorectal patients from the Darmkrebs: Chancen der Verhütung durch Screening (DACHS) study. For the classification models, the predicted dichotomised labels were used, whereas our regression model used the predicted continuous scores for the univariate Cox proportional-hazard models. An HR equal to 1 indicates non-significant prognostication capability.

|  | TCGA-CRC on DACHS-CRC | | | | TCGA-BRCA on DACHS-CRC | | | |
|---|---|---|---|---|---|---|---|---|
| Covariates/ CAMIL model | HR | 95%CI Low | 95%CI High | p-value | 95%CI Low | 95%CI High | P-value | p-value |
| **Sex** | 1.01 | 0.85 | 1.21 | 8.74E-01 | 1.01 | 0.85 | 1.21 | 8.74E-01 |
| **Age** | 1.03 | 1.02 | 1.04 | **1.08E-11** | 1.03 | 1.02 | 1.04 | **1.08E-11** |
| **Stage** | 1.81 | 1.65 | 2.00 | **3.25E-33** | 1.81 | 1.65 | 2.00 | **3.25E-33** |
| **Classification – LF** | 0.83 | 0.70 | 0.99 | **3.94E-02** | 0.91 | 0.77 | 1.08 | 2.98E-01 |
| **Regression – LF** | 0.20 | 0.06 | 0.62 | **5.50E-03** | 0.24 | 0.10 | 0.54 | **6.66E-04** |
| **Classification – LISS** | 0.82 | 0.69 | 0.98 | **2.65E-02** | 0.95 | 0.80 | 1.12 | 5.38E-01 |
| **Regression – LISS** | 0.14 | 0.04 | 0.53 | **3.90E-03** | 0.28 | 0.11 | 0.68 | **5.08E-03** |
| **Classification – Prolif.** | 1.09 | 0.92 | 1.30 | 3.15E-01 | 1.25 | 1.06 | 1.49 | **9.44E-03** |
| **Regression – Prolif.** | 1.52 | 0.21 | 11.07 | 6.78E-01 | 4.16 | 1.61 | 10.80 | **3.35E-03** |
| **Classification – SF** | 0.87 | 0.73 | 1.04 | 1.31E-01 | 1.01 | 0.85 | 1.19 | 9.51E-01 |
| **Regression – SF** | 0.40 | 0.16 | 1.05 | 6.17E-02 | 0.28 | 0.13 | 0.61 | **1.46E-03** |
| **Classification – TIL RF** | 0.97 | 0.82 | 1.16 | 7.46E-01 | 1.13 | 0.95 | 1.34 | 1.74E-01 |
| **Regression – TIL RF** | 0.49 | 0.15 | 1.57 | 2.30E-01 | 0.33 | 0.09 | 1.16 | 8.41E-02 |

**Suppl. Table 13: Hazard ratios (HR) with 95% confidence interval (CI) and corresponding p-values of multivariate Cox proportional-hazard models for the models for biological process biomarkers.** The CAMIL regression and CAMIL classification models were trained using site-aware splits on cohorts from The Cancer Genome Atlas colorectal cancer (CRC) and breast cancer (BRCA) on biomarkers for tumor infiltrating lymphocytes regional fraction (TIL RF), proliferation (Prolif.), leukocyte fraction (LF), lymphocyte infiltration signature score (LISS), and stromal fraction (SF). These models were deployed on CRC patients from the Darmkrebs: Chancen der Verhütung durch Screening (DACHS) study. For the classification models, the predicted dichotomised labels were used, whereas our regression model used the predicted continuous scores for the univariate Cox proportional-hazard models. The covariates used in the analysis are sex, age, and tumor stage. An HR equal to 1 indicates non-significant prognostication capability.

|  | TCGA-CRC on DACHS-CRC | | | |
|---|---|---|---|---|
| CAMIL model | HR | 95%CI Low | 95%CI High | P-val |
| **Classification – SF** | 0.77 | 0.65 | 0.89 | 2.14E-05 |
| **Regression – SF** | 0.84 | 0.72 | 0.96 | 3.20E-03 |
| **Classification – LISS** | 0.74 | 0.62 | 0.86 | 1.35E-06 |
| **Regression – LISS** | 0.72 | 0.60 | 0.85 | 1.12E-06 |
| **Classification – LF** | 0.74 | 0.62 | 0.86 | 1.32E-06 |
| **Regression – LF** | 0.72 | 0.60 | 0.85 | 1.03E-06 |
| **Classification – Prolif.** | 1.05 | 0.93 | 1.17 | 3.90E-01 |
| **Regression – Prolif.** | 0.94 | 0.82 | 1.06 | 2.79E-01 |
| **Classification – TIL RF** | 0.91 | 0.79 | 1.03 | 1.20E-01 |
| **Regression – TIL RF** | 0.86 | 0.74 | 0.97 | 1.02E-02 |

**Suppl. Table 14: Hazard ratios (HR) with 95% confidence interval (CI) and corresponding p-values of univariate Cox proportional-hazard models for the models for biological process biomarkers with regression scores binarized at the median.** The regression and classification models were trained using site-aware splits on cohorts from The Cancer Genome Atlas colorectal cancer (CRC) and breast cancer (BRCA) on biomarkers for tumor infiltrating lymphocytes regional fraction (TIL RF), proliferation (Prolif.), leukocyte fraction (LF), lymphocyte infiltration signature score (LISS), and stromal fraction (SF). These models were deployed on colorectal patients from the Darmkrebs: Chancen der Verhütung durch Screening (DACHS) study. For both the classification and regression models, the predicted dichotomised labels were used. An HR equal to 1 indicates non-significant prognostication capability.

Data availability

|  | **TCGA-CRC** | **DACHS-CRC** |
|---|---|---|
| Usage | Model training | Overall Survival (OS) |
| Cohort type | Population | Population |
| # of patients | 632 | 2448 |
| Median OS (days) | - | 3604 [3355 - 3839] |
| Age (median) | 68 | 69 |
| Age (IQR) | 18 | 14 |
| Sex: Male | 322 (50.9%) | 1436 (58.7%) |
| Sex: Female | 292 (46.2%) | 1012 (41.3%) |
| Sex: Unknown | 18 (2.85%) | 0 |
| Stage 1 | 76 (12%) | 485 (19.8%) |
| Stage 2 | 166 (26.3%) | 801 (32.7%) |
| Stage 3 | 140 (22.2%) | 822 (33.6%) |
| Stage 4 | 63 (10%) | 337 (13.8%) |
| Stage unknown | 187 (29.5) | 3 (0.1%) |
| Left-sided CRC | 248 (39.2%) | 1607 (65.6%) |
| Right-sided CRC | 176 (27.8%) | 819 (33.5%) |
| Unknown side | 209 (33%) | 22 (0.9%) |

**Suppl. Table 15: Clinical features of The Cancer Genome Atlas (TCGA) and the Darmkrebs: Chancen der Verhütung durch Screening (DACHS) colorectal cancer (CRC) cohort.** The external cohort with CRC patients for the biological process biomarkers came from the DACHS study. The TCGA cohort is utilized for model training, whereas the DACHS cohort is utilized for overall survival (OS) prediction in CRC. The median OS of the DACHS cohort, expressed in days, consists of the 0.95 lower confidence level and the 0.95 higher confidence level. The TCGA-CRC cohort has clinical information for 632 patients, but only whole-slide images for 625 patients are available. The DACHS-CRC cohort has clinical information for 2448 patients, but only 2297 have overlapping whole-slide images and corresponding survival data.

| **HRD** | | | | | | | **3273** | | | | **452** |
|---|---|---|---|---|---|---|---|---|---|---|---|
| Cohort | TCGA | | | | | | | CPTAC | | | |
| Cancer types | BRCA | UCEC | PAAD | CRC | LUAD | GBM | LUSC | UCEC | PAAD | LUSC | LUAD |
| N slides | 1133 | 566 | 209 | 625 | 544 | 860 | 512 | 883 | 557 | 1081 | 1137 |
| N features | 1133 | 566 | 209 | 599 | 529 | 860 | 512 | 883 | 557 | 1081 | 1125 |
| N target overlap | 1005 | 467 | 173 | 496 | 449 | 232 | 451 | 99 | 139 | 108 | 106 |
| N HRD+ | 281 | 68 | 13 | 16 | 158 | 6 | 232 | 3 | 4 | 33 | 14 |
| N HRD- | 724 | 399 | 160 | 480 | 291 | 226 | 219 | 96 | 135 | 75 | 92 |

**Suppl. Table 16: Data availability for the homologous recombination deficiency (HRD) target.** Data availability of patients from The Cancer Genome Atlas (TCGA), breast cancer (BRCA), colorectal cancer (CRC), glioblastoma (GBM), lung adenocarcinoma (LUAD), lung squamous cell cancer (LUSC), pancreatic cancer (PAAD) and endometrial cancer (UCEC) for HRD. The external

cohorts for HRD were from the Clinical Proteomic Tumor Analysis Consortium (CPTAC) effort, consisting of UCEC, PAAD, LUSC and LUAD.

| **LISS** | | | | | | | **3636** | **0** |
|---|---|---|---|---|---|---|---|---|
| Cohort | TCGA | | | | | | | DACHS |
| Cancer types | BRCA | UCEC | STAD | CRC | LUAD | LIHC | LUSC | CRC |
| N slides | 1133 | 566 | 209 | 625 | 544 | 860 | 512 | 3617 |
| N features | 1133 | 566 | 209 | 599 | 529 | 860 | 512 | 2297 |
| N target overlap | 1048 | 490 | 334 | 560 | 410 | 331 | 463 | 2297 |
| **SF** | | | | | | | **3513** | **0** |
| Cohort | TCGA | | | | | | | DACHS |
| Cancer types | BRCA | UCEC | STAD | CRC | LUAD | LIHC | LUSC | CRC |
| N slides | 1133 | 566 | 209 | 625 | 544 | 860 | 512 | 3617 |
| N features | 1133 | 566 | 209 | 599 | 529 | 860 | 512 | 2297 |
| N target overlap | 989 | 456 | 360 | 500 | 441 | 320 | 447 | 2297 |
| **TIL RF** | | | | | | | **3124** | **0** |
| Cohort | TCGA | | | | | | | DACHS |
| Cancer types | BRCA | UCEC | STAD | CRC | LUAD | LIHC | LUSC | CRC |
| N slides | 1133 | 566 | 209 | 625 | 544 | 860 | 512 | 3617 |
| N features | 1133 | 566 | 209 | 599 | 529 | 860 | 512 | 2297 |
| N target overlap | 943 | 447 | 335 | 555 | 459 | 0 | 385 | 2297 |
| **LF** | | | | | | | **3719** | **0** |
| Cohort | TCGA | | | | | | | DACHS |
| Cancer types | BRCA | UCEC | STAD | CRC | LUAD | LIHC | LUSC | CRC |
| N slides | 1133 | 566 | 209 | 625 | 544 | 860 | 512 | 3617 |
| N features | 1133 | 566 | 209 | 599 | 529 | 860 | 512 | 2297 |
| N target overlap | 1035 | 493 | 373 | 561 | 459 | 333 | 465 | 2297 |
| **Prolif.** | | | | | | | **3636** | **0** |
| Cohort | TCGA | | | | | | | DACHS |
| Cancer types | BRCA | UCEC | STAD | CRC | LUAD | LIHC | LUSC | CRC |
| N slides | 1133 | 566 | 209 | 625 | 544 | 860 | 512 | 3617 |
| N features | 1133 | 566 | 209 | 599 | 529 | 860 | 512 | 2297 |
| N target overlap | 1048 | 490 | 334 | 560 | 410 | 331 | 463 | 2297 |

**Suppl. Table 17: Data availability for the biomarkers related to biological processes.** Data availability of patients from The Cancer Genome Atlas (TCGA), breast cancer (BRCA), colorectal cancer (CRC), liver hepatocellular carcinoma (LIHC), lung adenocarcinoma (LUAD), lung squamous cell cancer (LUSC), gastric cancer (STAD) and endometrial cancer (UCEC) for biological process biomarkers: tumor infiltrating lymphocytes regional fraction (TIL RF), proliferation (Prolif.), leukocyte fraction (LF), lymphocyte infiltration signature score (LISS), and stromal fraction (SF). The external cohort with CRC patients for the biological process biomarkers came from the Darmkrebs: Chancen der Verhütung durch Screening (DACHS) study.